\newcommand{\framework}{HoLLMwood}
\def\paperTitle{
\textsc{\framework:} Unleashing the Creativity of Large Language Models in Screenwriting via Role Playing
}

\newcommand{\nbf}[1]{{\noindent \textbf{#1}}}
\newcommand*{\affmark}[1][*]{\textsuperscript{#1}}

\def\authorBlock{
    Jing Chen$^1$\footnotemark[1] \qquad
    Xinyu Zhu$^3$\footnotemark[1] \qquad
    Cheng Yang$^3$\footnotemark[3] \qquad
    Chufan Shi$^3$\footnotemark[3]  \\
    \textbf{
    Yadong Xi$^2$ \quad  
    Yuxiang Zhang$^4$ \quad
    Junjie Wang$^4$ \quad
    Jiashu Pu$^2$}
    \\
    \textbf{
    Rongsheng Zhang$^2$\footnotemark[2] \qquad
    Yujiu Yang$^3$ \qquad
    Tian Feng$^1$\footnotemark[2]}
    \\
    \affmark[1]Zhejiang University \qquad
    \affmark[2]Fuxi AI Lab, NetEase Inc.\\
    \affmark[3]Tsinghua University \qquad
    \affmark[4]Waseda University \\
    {\tt\small \{chenjing\_1984, t.feng\}@zju.edu.cn} \qquad
    {\tt\small zhangrongsheng@corp.netease.com} \\

}

\newif\ifreview 
\newif\ifarxiv \newcommand{\arxiv}{\arxivtrue}
\newif\ifcamera 
\arxiv

\documentclass[11pt]{article}
\usepackage{xcolor}
\definecolor{mycolor}{rgb}{0.9373, 0.8039, 0.8588}

\ifreview \usepackage[review]{ACL2023} \fi
\ifarxiv \usepackage{ACL2023} \fi
\ifcamera \usepackage{ACL2023} \fi

\usepackage{times}
\usepackage{latexsym}
\usepackage[T1]{fontenc}
\usepackage[utf8]{inputenc}
\usepackage{microtype}
\usepackage{inconsolata}
\usepackage{mdframed}
\usepackage{multirow}

\usepackage{graphicx}
\usepackage{amsmath}
\usepackage{amssymb}
\usepackage{booktabs}
\usepackage{algorithm}
\usepackage{algorithmic}
\usepackage{footnote}
\usepackage[font=small,labelfont=bf]{caption}

\usepackage{array}
\usepackage{longtable}
\usepackage{enumitem}
\usepackage{xcolor}
\usepackage{colortbl}
\usepackage{comment}
\usepackage{amsthm}
\usepackage{bbm} % indicator function
\usepackage{subcaption}
\usepackage{multirow}
\usepackage{comment}
\usepackage{makecell} %multiline cell
\usepackage{multicol}
\usepackage[normalem]{ulem}
\useunder{\uline}{\ul}{}
\usepackage{booktabs}
\usepackage[switch]{lineno}

\usepackage{url}
\usepackage{xspace}

\usepackage[export]{adjustbox}
\usepackage{paralist} 

\usepackage{tabu}
\usepackage{tikz}

\newcommand{\eat}[1]{}

\newcommand{\cbit}{\begin{compactitem}}
\newcommand{\ceit}{\end{compactitem}}
\newcommand{\cben}{\begin{compactenum}}
\newcommand{\ceen}{\end{compactenum}}

\definecolor{shadecolor}{rgb}{0.9,0.9,0.9}

\definecolor{headercolor}{rgb}{0.9,0.9,0.9}
\definecolor{headercolor}{RGB}{240, 255, 240}

\definecolor{rowcolor}{rgb}{0.8, 0.85, 0.88}
\definecolor{rowcolor1}{rgb}{1.0, 0.97, 0.91} %
\definecolor{rowcolor1}{RGB}{255, 240, 240} %

\definecolor{rowcolor2}{rgb}{0.95, 1, 1} %
\definecolor{rowcolor3}{rgb}{0.835, 0.792, 0.9215} %
\definecolor{rowcolor2}{rgb}{0.835, 0.85, 0.96}
\definecolor{rowcolor2}{RGB}{213, 217, 245}
\definecolor{rowcolor2}{RGB}{240, 240, 255}

\definecolor{rowcolor3}{rgb}{1,1,1}

\definecolor{highlightcolor}{rgb}{1, 0, 0} %
\definecolor{keybgcolor}{rgb}{1, 1, 0.8} %

\newcommand\keyword[1]{%
\tikz[baseline=(word.base)]{
  \node[rounded corners=2pt,fill=rowcolor,anchor=base] (word) {#1};}%
}

\mdfdefinestyle{MyShadeQuoteStyle}{%
    leftmargin=15pt,
    rightmargin=15pt,
    backgroundcolor=gray!25,
    linewidth=0pt,
    skipbelow=\topskip,
    skipabove=\topskip
}

\usepackage[capitalize]{cleveref}
\crefname{section}{Sec.}{Secs.}
\crefname{table}{Table}{Tables}
\crefname{figure}{Fig.}{Figs.}

\usepackage{tabularx}

\title{\paperTitle}
\author{\authorBlock}

\begin{document}
\maketitle

{
  \renewcommand{\thefootnote}%
  {\fnsymbol{footnote}}
  \footnotetext[1]{Equal contribution.}
    \footnotetext[3]{Co-second authors.}
  \footnotetext[2]{Corresponding authors.}

}

\begin{abstract}
Generative AI has demonstrated unprecedented creativity in artistic creation, especially in the field of computer vision, yet such phenomena have not been observed in natural language processing. In particular, large language models (LLMs) can hardly produce written works at the level of human experts due to the extremely high complexity of literature writing. 
In this paper, we present \textsc{\framework}, an automated framework for unleashing the creativity of LLMs and exploring their potential in screenwriting, which is a highly demanding task.
Mimicking the human creative process, we assign LLMs to different roles involved in the real-world scenario. In addition to the common practice of treating LLMs as \emph{Writer}, we also apply LLMs as \emph{Editor}, who is responsible for providing feedback and revision advice to \emph{Writer}. Besides, to enrich the characters and deepen the plots, we introduce a role-playing mechanism and adopt LLMs as \emph{Actors} that can communicate and interact with each other. Evaluations on automatically generated screenplays show that \textsc{\framework} substantially outperforms strong baselines in terms of coherence, relevance, interestingness and overall quality.
\end{abstract}

\section{Introduction}
\label{sec:introduction}

Artistic creation, which has long been regarded as the unique intelligence of human beings, is being redefined by the recent advancements in artificial intelligence. The latest generative models have shown notable creativity, especially in the field of computer vision \citep{dalle3, diffusion}. 
These AI-generated artworks can sometimes achieve a quality indistinguishable from human-created art, blurring the conventional boundaries between human and machine-created arts.
However, even those generative models are capable of creating excellent visual artworks on par with those crafted by human experts, the same level of achievements have not yet been observed in the realm of literary creation.

Screenwriting, the art and craft of writing screenplays, serves as the cornerstone in the production of films, TV series and video games. This highly creative process demands not only deliberate thinking and careful planning, but also an in-depth understanding of human emotions and motivations. 
\begin{figure}[t]
    \centering
    \includegraphics[width=0.98\linewidth]{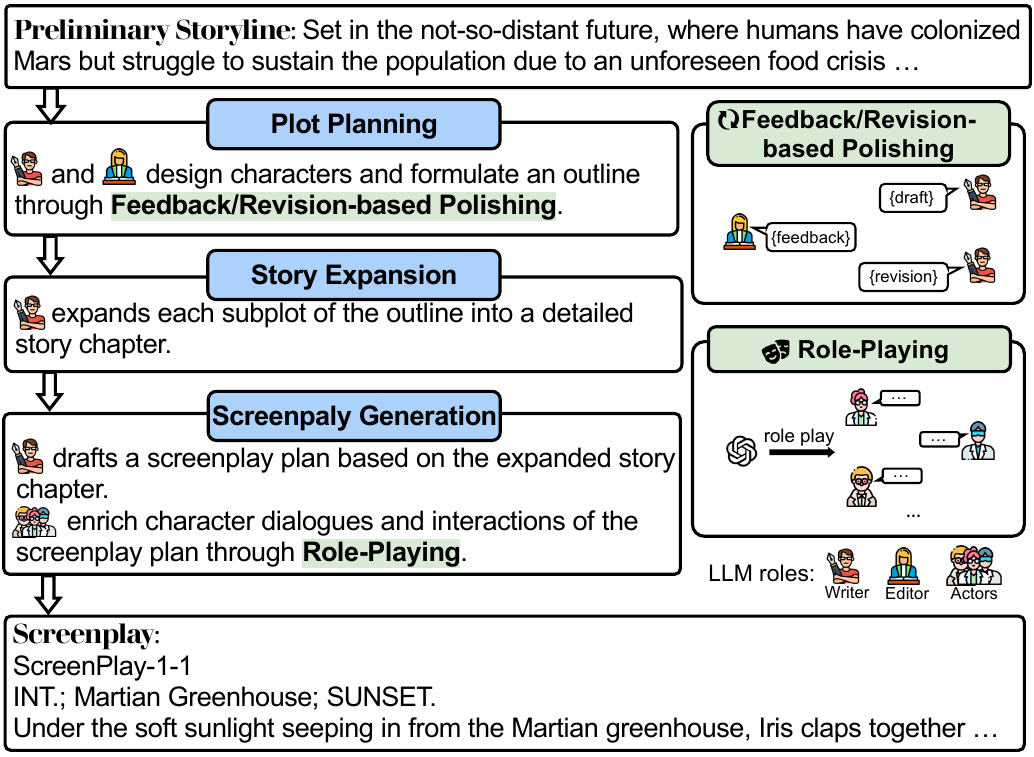}
    \caption{
    An overview of \textsc{\framework} for automatic screenwriting.
    }
    \label{fig:introduction}
\end{figure}
Despite the recent advances in large language models (LLMs) \citep{DBLP:conf/nips/BrownMRSKDNSSAA20@gpt3, palm-google, chinchilla}, it remains extremely challenging for current LLMs to produce high-quality literary works with simple guidelines. Some prior works have explored employing LLMs for such creative tasks, e.g., story generation \citep{re3, DOC, yuan2022wordcraft} and co-writing scripts with human experts \citep{screenwriting-deepmind}, while either dialogues in the story are clearly robotic, or human expertise is required, which comes at a high cost. Recent works have explored the potential of solving real-world problems in a multi-agent setting with LLMs~\cite{Gravitas2023, AntonOsika2023, yoheinakajima2023}, while little progress has achieved in creative writing under such setting.
This can be attributed to the intricate nature of literary creation. An engaging work requires authors to constantly polish the drafts based on external feedback, step into the shoes of their characters, experience the narrative from their perspectives, and cultivate a deep emotional connection with the audience.
Therefore, simple guideline may fail in making LLMs write satisfactory works.

In order to explore LLMs' potential and capability in creative writing, we focus on the aforementioned screenwriting task, since it's close to life and has promising and valuable applications. 
We design a fully automated screenwriting framework, named \textsc{\framework}, aiming to make LLMs mimic the creative process of Hollywood screenwriters.
Specifically, we organize LLMs into different roles involved in the human creative process: \textit{Writer} and \textit{Editor}. Editor will provide revision advice for writer on characters and plots. Additionally, to bring the characters in the plot to life and flesh them out, we make LLMs act as \textit{Actors} and develop dialogues and interactions between characters in a role-playing manner. Each step of the screenwriting process in our framework is separated from each other, making it easy to introduce human intervention at any stage. % 

We evaluate the generated screenplays via pairwise comparison using GPT-4. The experiments show that the scripts generated with \textsc{\framework} achieve significantly overall better quality compared to strong baselines. We also perform more fine-grained evaluations of different dimensions (coherence, relevance and interestingness). Ablations further prove that both feedback-revision between writer and editor, and role-playing by actors contribute positively to the final screenplay quality.

In summary, our contributions are as follows:
\begin{itemize}
    \item Experimental results reveal that LLMs fall short in producing literary works of high quality under simple guidelines. Specifically, they struggle to generate screenplays featuring vivid characters and engaging plots. The dialogues and interactions among characters often appear robotic and boring, indicating the challenges in directly applying LLMs for creative tasks.  
    \item We propose a fully automated framework for screenwriting with LLMs, named \textsc{\framework}. With our framework, not only the non-expert users can create engaging screenplays, but also the industry professionals will have an assistant to draft new ideas. Users only need to provide a preliminary storyline as input and our approach handles the intricate task automatically, thus democratizing a field traditionally reserved for those with extensive experience or specific skills.
    \item  Pairwise comparison using GPT-4 shows the superiority of \textsc{\framework} compared to other prompting methods on synthesized storylines. Ablations further prove the effectiveness of each module in our framework.
\end{itemize}

\section{\textsc{\framework}}
\label{sec:method}
In this section, we present the \textsc{\framework} framework that enables LLMs to create engaging screenplays. The LLMs will play three main roles in the whole screenwriting process: \emph{Writer}, \emph{Editor} and \emph{Actors} under carefully designed prompts.

In the beginning, the writer designs a set of characters and formulates a story's outline (\cref{outline_and_characters}), grounded in a preliminary storyline. The editor will provide some feedbacks for the writer to polish the initial draft iteratively. Then, the subplots within the outline are further elaborated by the writer, each of which is expanded into a full chapter of the story (\cref{expand_outline}). 
Building upon this, the writer drafts a screenplay that includes scenes, characters, actions and conversations, with each tailored to the corresponding chapter (\cref{script_generation}). 
To deepen the character performances and elevate the overall interestingness, we introduce a role-playing mechanism to develop the final screenplay, where the LLMs are assigned specific characters and prompted to act spontaneously based on the current plot, characters' profile and memory.
Complete examples of inputs and outputs for \textsc{\framework}'s each step are provided in \cref{append:case_pre}.

\subsection{Plot Planning}
\label{outline_and_characters}

Given a preliminary storyline, \textsc{\framework} starts with designing characters and formulating a story's outline containing a set of plots.
\begin{figure}[!htbp]
\centering
\includegraphics[width=0.98\linewidth]{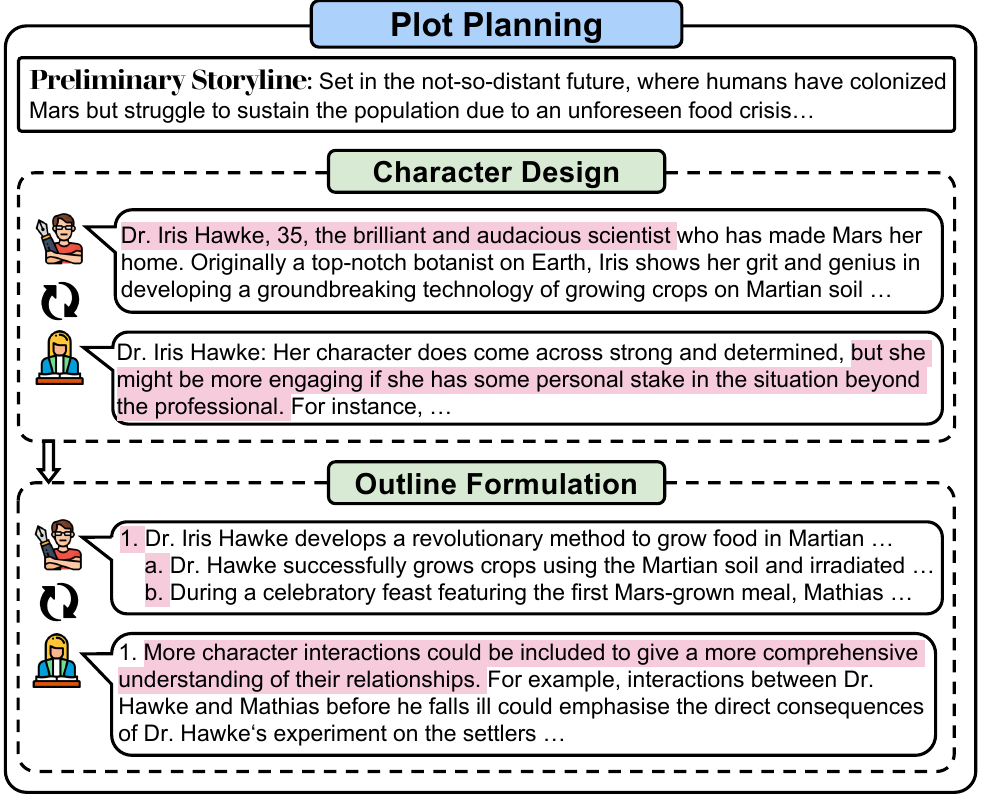}
\caption{An example of plot planning. \colorbox{mycolor}{Highlighted texts} refer to the feedback parts proposed by the editor.}
    \label{fig:plot_planing}
\end{figure}

\nbf{Character Design.}
As characters are the soul of the plots, the writer first designs a certain number of characters based on the preliminary storyline. Each character has a \emph{name} and an \emph{introduction} as illustrated in \cref{fig:plot_planing}. In our implementation, the number of characters in the story ranges from $3$ to $6$, since extremely limited characters would lead to monotonous relationships whereas too many characters might weaken their interactions.
The detailed prompt used to generate characters is presented in appendix \cref{tab:char_gen_prompt}.  

\nbf{Outline Formulation.} Taking the designed characters and preliminary storyline as input, the writer is prompted to formulate a complete outline for the story. 
We structure the outline in a hierarchical manner similar to DOC~\citep{DOC}. However, we find a two-level outline capable to hold adequate plots. Therefore, we adopt a two-level hierarchy in our framework with multiple subplots below the top tier of plots, rather than the three-level setting in DOC.  Each subplot comprises scene, involved characters and storyline. The detailed prompt used to generate outline is given in \cref{tab:out_gen_prompt}.

Unlike prior works, which typically generate the outline in multiple steps, we generate the whole outline in a single pass. To maintain a clear structure and facilitate parsing of long context, we employ an HTML-style prompting format, using ``<\textit{component}>'' and ``</\textit{component}>'' to wrap the different components. A detailed example is shown in appendix \cref{tab:output_format}. 

\nbf{Feedback/Revision-based Polishing.} 
In the real world, human writers would likely keep polishing a draft manuscript until it meets a certain expectation or quality, which usually involves communications with others, especially editors. Our framework mimics this process via applying an LLM as an editor, who is prompted to provide feedback for the writer to revise the initial drafts.
Specifically, the editor is prompted to provide suggestions that focus on the following aspects:
\begin{itemize}
    \item Do the characters align satisfactorily with the preliminary storyline?
    \item Are the characters portrayed engaging and emotionally resonant to the audience?
    \item Does the outline conflict with the preliminary storyline and the characters?
    \item Are the plots in the outline consistent and coherent?
    \item Are the plots able to form an interesting, engaging, and thought-provoking story?
\end{itemize}
\begin{figure}[!htbp]
\centering
\includegraphics[width=0.98\linewidth]{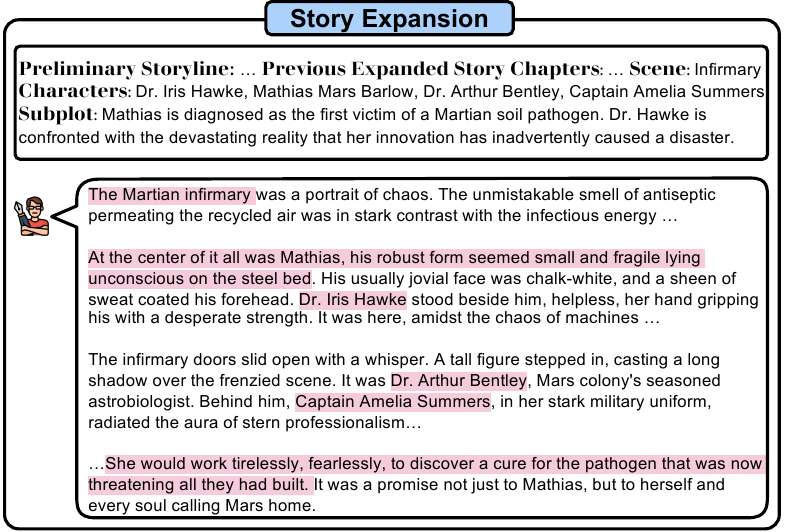}
\caption{An example of story expansion. \colorbox{mycolor}{Highlighted texts} refer to the parts expanded from the given input.}

    \label{fig:story_expansion}
\end{figure}

Besides, the editor is configured to trace whether the story has a clear ending and provide advice if necessary. 
We adopt an iterative polishing strategy for character design and outline formulation, following the real-world scenario where a draft needs repeated revisions to reach excellence.
We set the maximum iteration to $2$ and the editor is prompted to stop providing feedback when it considers the draft satisfactory.
The detailed prompts used to guide interaction between the writer and editor is presented in appendix \cref{tab:advice_char_prompt,tab:advice_out_prompt,tab:revise_prompt,tab:advice_again_prompt}.

\begin{figure*}[!tp]
    \centering
    \includegraphics[width=\textwidth]{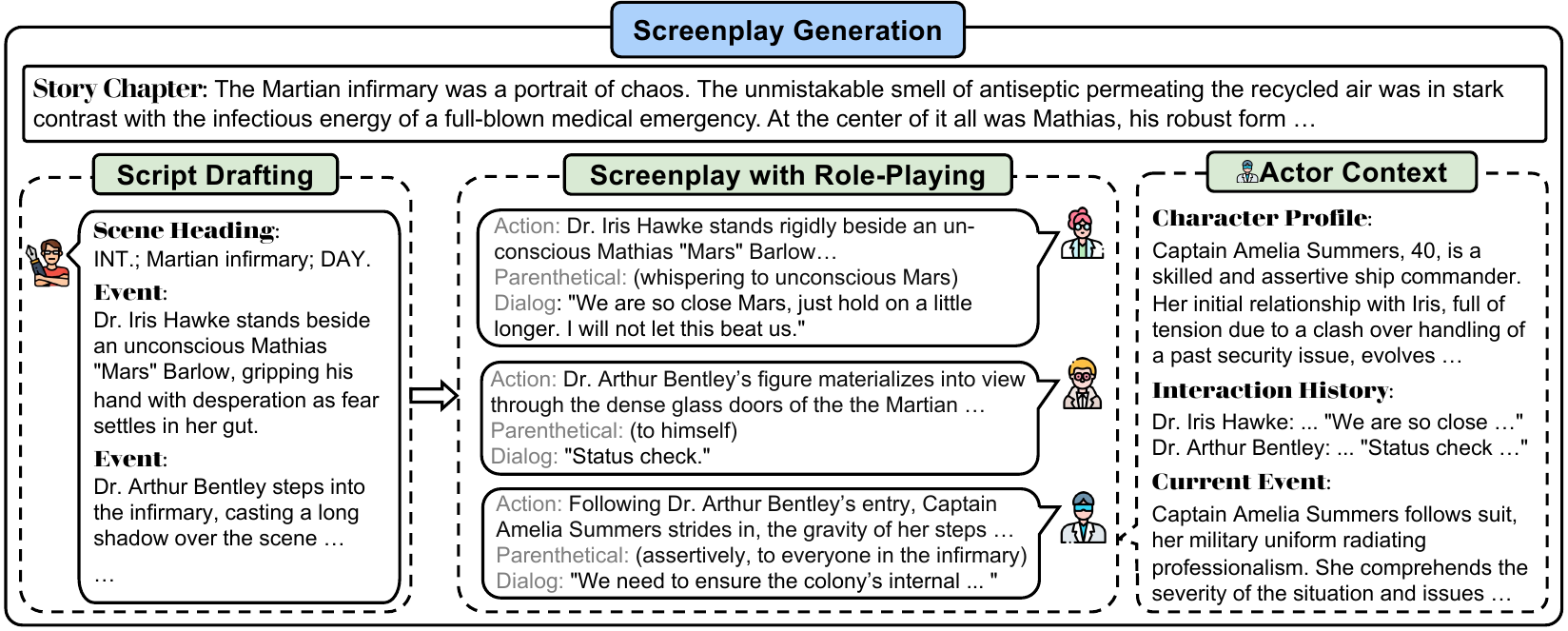}
    \caption{
An example of screenplay generation with the script draft in the left, the screenplay with role playing in the middle and the actor context in the right.
    }
    \label{fig:role_playing}
\end{figure*}

\subsection{Story Expansion}
\label{expand_outline}
After carefully designing the characters and outline, we move on to adding more details to the plots, while maintaining coherence between adjacent plots and relevance to the initial storyline.

\nbf{Coherence Maintaining.}
The writer is prompted to elaborate each subplot and expand it into a full chapter of the story.
To maintain coherence throughout the story, it is crucial to incorporate what happened before into the current expansion. To achieve this, we include the nearest $n$ chapters preceding the current subplot as context when expanding it. Chapters situated at a distance greater than $n$ will not be considered, instead, we include the unexpanded subplots. We set $n$ to 1 in the experiments. 
Additionally, the plot, scene, and introductions of involved characters in the current subplot will always be put in the beginning of the context.
See \cref{tab:expand_prompt} in appendix for the detailed prompt used for story expansion.

\subsection{Screenplay Generation}
\label{script_generation}
In the entertainment and film industry, it is quite common to adapt novels or other types of literary works into TV series and movies. \textsc{\framework} simulates this practice in the real world and adapts the story chapters into an engaging screenplay that unfolds over several episodes.

\nbf{Script Format.} % 
Typically, an episode script consists of several key elements that make up a blueprint for the production of a film:
\begin{enumerate}
    \item \textit{Scene Heading.} Describe the location and time of day for a particular scene.
    \item \textit{Action Line.} Describe the actions and events occurring in the scene. 
    \item \textit{Character Name.} Indicate who is the main character shown on the screen.
    \item \textit{Dialogue.} The spoken words of characters, it is the central element that conveys the story. % 
    \item \textit{Parenthetical.} Information used to provide additional instructions about how a line of dialogue should be delivered. %
\end{enumerate}
These elements are connected in the format of ``Scene Heading ... Character Name: [Action Line] (Parenthetical) Dialogue ...'' to form a script, as shown in the right of \cref{tab:draft_screenplay}.

\nbf{Script Drafting.} 
Taking the chapters expanded from subplots, the writer drafts an initial screenplay, with each chapter converted into an episode of script draft as shown in the left of \cref{fig:role_playing}. 
The script draft consists of two main components: the Scene Heading and a series of events that happen sequentially. Each event briefly describes a single character's behavior, such as ``Tom walks into the cafe and orders a cup of coffee'', which will provide instructions for subsequent role-playing. 
To fit the narrative of movies and TV series, we prompt the writer to make appropriate cuts and adjustments, as a chapter usually contains redundant descriptions of environments and characters' psychological activities.
See \cref{tab:draft_prompt} in appendix for the detailed prompt used for script drafting.

\nbf{Role-Playing.} Instead of directly generating character interactions from a third-person perspective, we aim to bring the characters to life by casting LLMs in the roles of those characters.
Building on the script draft generated before, we position LLMs as the characters and put them into the plots to interact with each other from a first-person perspective as shown in the middle of \cref{fig:role_playing}.

The role-playing is carried out sequentially in accordance with the events outlined in the script draft. Since the draft only sketches a broad overview of the performance, specific details are left to the actors for creative interpretation. 
Specifically, we initialize each actor with the introduction of the corresponding character and include its interaction history with other characters as context. 
The instantiated characters are then immersed in the current scene sandbox, spontaneously delivering performances grounded in its personality and the current event. % 
Each actor's performance will include Action Line, Parenthetical, and Dialogue as mentioned above. Therefore, the development of the plots, as well as the dialogues between characters, unfold in a dynamic, role-playing manner.
The actors' detailed interpretation of all events forms the script for an episode.
All episodes together comprise a complete screenplay as the final output.
The detailed prompt for role-playing is presented in \cref{tab:act_prompt}.

\setlist[enumerate]{topsep=0pt,itemsep=-1ex,partopsep=1ex,parsep=1ex}

\section{Experiments}
\label{sec:experiments}

\subsection{Experimental Setup}
\label{exp_setup}
\subsubsection{Dataset}
\label{dataset}
We use LLMs to synthesize preliminary storylines across several common film genres as input as input. Specifically, we include six different genres: \textit{Romance, Science Fiction, Horror, Drama, Crime, Comedy}. For each genre, we create 10 examples with \texttt{gpt-4-0613}, resulting in 60 instances for testing. To facilitate pairwise comparison, we set the top-level plots in the outline the same for baselines. Each storyline contains approximately 120 words. Detailed information on the synthesized dataset is given in ~\cref{append:dataset_details}.

\subsubsection{Baselines}
\label{baselines}
We find that there are limited existing works focused on applying LLMs for automatic screenwriting, 
but only methods on storytelling with LLMs. We make some adjustments to these methods to generate screenplays that can be compared with ours. 
We use the following two baselines:

\begin{enumerate}

\item {Plan-then-Write}, we prompt LLMs to design several characters and formulate an outline given the storyline, and then create each episode's script sequentially according to the plots within the outline.

\item {DOC-screen}, 
we use DOC \citep{DOC} to generate chapters of the story and then prompt LLMs to generate scripts based on each chapter.
\end{enumerate}
We provide LLMs with a single-episode script as an in-context learning example for both baselines. 
We use \texttt{gpt-3.5-turbo-16k-0613} and \texttt{gpt-4-0613} for all the methods, \texttt{gpt-4-32k-0613} for evaluation. For hyperparameters, we set temperature to 1 and top\_p sampling to 0.999.

\subsubsection{Evaluation}
\label{eval}
It's typically difficult to quantitatively assess the results in creative writing tasks with automatic metrics \citep{re3, DOC}.
In our evaluation, we conduct pairwise comparison between the screenplays generated by different methods using GPT-4.
Considering a full screenplay is too long (>5000 words), we segment it based on top-level plots, resulting in 212 and 206 pairs generated with GPT-3.5 and GPT-4, respectively. 
For each comparison, we concatenate scripts belonging to the same top-level plot (approximately 1500 words) and conduct pairwise comparisons between examples generated with our method and baselines.
We evaluate a screenplay in four dimensions: 
\begin{enumerate}
    \item \textit{Coherence.} 
    Evaluate the coherence from the plot structure, character description, scene transitions and setup consistency.
    \item \textit{Relevance.} Evaluate the relevance from the relationship between the top-level plot and the final scripts.
    \item \textit{Interestingness.} Evaluate the interestingness from the originality of the screenplay, the unexpectedness of plots, the depth of the characters, and the vividness of the dialog.
    \item \textit{Overall Quality.} Consider the coherence, relevance, and interestingness of a screenplay to assess the overall quality.
\end{enumerate}
For each metric, we require GPT-4 to decide which screenplay is better, or the two are indistinguishable from each other (`tie'). 
For the detailed evaluation prompt, see \cref{tab:eval_prompt,tab:eval_focuses} in appendix.

\begin{table*}[htbp]
\small
% \normalsize
\centering
\begin{tabular}{cc|cccc}
\toprule
\textbf{Backbone} & \textbf{Method} & \textbf{Coherence }$\uparrow$ & \textbf{Relevance} $\uparrow$ & \textbf{Interesting} $\uparrow$ & \textbf{Overall} $\uparrow$ \\

\midrule
\multirow{6}{*}{\texttt{gpt-3.5-turbo}}& \textsc{Plan-then-Write} Wins & 43.2     & 41.2     &   38.8    & 41.3         \\
& \textit{\textsc{\framework}} Wins & \textbf{56.8} & \textbf{57.8}       & \textbf{60.2}      & \textbf{57.8}          \\
& Ties & {0.0} & {1.0}      & {1.0}      & {1.0}          \\
\cmidrule{2-6}
& \textsc{DOC-screen} Wins & 45.6 & 42.7      & 42.7      & 43.2          \\
& \textit{\textsc{\framework}} Wins & {54.4} & \textbf{57.3}       & \textbf{56.8}      & \textbf{56.8}          \\
& Ties & {0.0} & {0.0}       & {0.5}      & {0.0}          \\
\midrule
\multirow{6}{*}{\texttt{gpt-4}} & \textsc{Plan-then-Write} Wins & 23.6 & 30.7       & 15.6        & 16.5           \\
& \textsc{\framework} Wins & \textbf{76.4} & \textbf{68.4}      & \textbf{84.0}      & \textbf{83.0}          \\
& Ties & {0.0} & {0.9}       & {0.4}      & {0.5}          \\
\cmidrule{2-6}
& \textsc{DOC-screen} Wins & 29.7 & 38.7       & 16.5      & 20.8          \\
& \textit{\textsc{\framework}} Wins & \textbf{70.3} & \textbf{60.8}       & \textbf{81.6}      & \textbf{79.2}          \\
& Ties & {0.0} & {0.5}      & {1.9}      & {0.0}          \\
\bottomrule
\end{tabular}
\caption{
Pairwise comparison between \textsc{\framework} and baselines using GPT-4 as the judge model. Evaluation is conducted separately on the four dimensions.
\textbf{Bold} indicates significance with p < 0.05.
}
\label{tab:results_baselines}
\vspace{-0.5em}
\end{table*}

\subsection{Results}
\label{sec:results}
As shown in \cref{tab:results_baselines}, \textsc{\framework} are recognized substantially better than all baselines. Plan-then-Write gets the lowest win rate when compared with \textsc{\framework}, which indicates that LLMs fall short in creative tasks, such as screenwriting, without meticulous guidelines.
DOC-screen is slightly better in comparison, while it still underperforms our framework, as it has no module carefully designed for generating stories with rich character performances and interactions.

Screenplays generated with \textsc{\framework} are preferred across four dimensions using both \texttt{gpt-3.5-turbo} and \texttt{gpt-4} as backbones. 
Interestingly, we find that \textsc{\framework} shows particularly better performance in interestingness and overall quality compared to coherence and relevance. This is expected, as the baselines lack a role-playing mechanism that allows LLMs to spontaneously act as live characters and interact with others.
We also find that \texttt{gpt-4} generally achieves higher win rate compared to \texttt{gpt-3.5-turbo} as backbone, this indicates that stronger model can benefit more from our framework. Consequently, our approach can grow alongside the increasing capabilities of foundation models and hopefully generate screenplays close to the level of humans.

\begin{table}[tp]
\centering
\small
\begin{tabular}{cc|cccc}
\toprule
\textbf{Content} & \textbf{Method} & \textbf{Co}$\uparrow$ & \textbf{Re}$\uparrow$ & \textbf{In}$\uparrow$ & \textbf{OQ}$\uparrow$ \\
\midrule
\multirow{6}{*}{\textbf{Chars}} &  \textit{R1} Wins & \textbf{86.7} & \textbf{83.3} & \textbf{88.3} & \textbf{85.0} \\
 & \textit{R0} Wins & 11.7 & 15.0 & 10.0 & 13.3 \\
 & Ties & 1.6 & 1.7 & 1.7 & 1.7 \\
\cmidrule{2-6}
 & \textit{R2} Wins & \textbf{88.3} & \textbf{80.0} & \textbf{91.7} & \textbf{88.3} \\
 & \textit{R0} Wins & 10.0 & 16.7 & 6.7 & 8.3 \\
 & Ties & 1.7 & 3.3 & 1.6 & 3.4 \\
\midrule
\multirow{6}{*}{\textbf{Outline}} &  \textit{R1} Wins & \textbf{96.7} & \textbf{90.0} & \textbf{98.3} & \textbf{93.3} \\
 & \textit{R0} Wins & 3.3 & 10.0 & {1.7} & 5.0 \\
 & Ties & 0.0 & 0.0 & 0.0 & 1.7 \\
\cmidrule{2-6}
 & \textit{R2} Wins & \textbf{96.7} & \textbf{91.7} & \textbf{98.3} & \textbf{95.0} \\
 & \textit{R0} Wins & 3.3 & 8.3 & 1.7 & 5.0 \\
 & Ties & 0.0 & 0.0 & 0.0 & 0.0 \\
\bottomrule 
\end{tabular}
\caption{\small Comparisons of characters and outlines under different number of feedback rounds. 
\textbf{Co}, \textbf{Re}, \textbf{In} and \textbf{OQ} are short for the aforementioned four dimensions, respectively. 
`\textit{R0}'(`\textit{R1}',`\textit{R2}') means the maximum feedback rounds are 0(1,2). 
% The `maximum' (e.g., \textit{n}) means that the Editor will provide up to \textit{n} rounds of feedback, but it can stop earlier when finding there is nothing to improve.
\textbf{Bold} indicates significance with p < 0.05.}
% With-feedback overwhelmingly beats without-feedback.}
\label{table:abla1_2}
\end{table}

\section{Ablation Study}
\label{sec:ablation}
\subsection{Impact of Feedback-Revision}
In this section, we investigate the impact of the feedback-revision mechanism on plot-planning.
Specifically, we compare the character decriptions and the outlines generated with different feedback rounds using GPT-4 as backbone. The evaluation is conducted in the same way as the main experiments.
As illustrated in \cref{table:abla1_2}, contents generated with feedback overwhelmingly beat those without feedback, and the win rate increases accordingly with more rounds of feedback.
Surprisingly, the feedback-revision not only contributes positively to the coherence and relevance, but also to the interestingness and overall quality, indicating that the feedback-revision can facilitate the following script development.

\begin{table}[tp]
\centering
\small
\begin{tabular}{c|c|cccc}
\toprule
\textbf{Backbone} & \textbf{Method} & \textbf{Co }$\uparrow$ & \textbf{Re}$\uparrow$ & \textbf{In}$\uparrow$ & \textbf{OQ}$\uparrow$ \\
\multirow{3}{*}{\texttt{gpt-3.5}} & {{w/  RP}} Wins & \textbf{89.0} & \textbf{79.5} & \textbf{89.2} & \textbf{84.3} \\
& {{w/o RP}} Wins & 10.5  & 20.3  & 8.6 & 14.5 \\
& Ties & 0.4  & 0.2  & 2.2 & 1.1 \\
\cmidrule{1-6}
\multirow{3}{*}{\texttt{gpt-4}} & {{w/  RP}} Wins & \textbf{76.9} & \textbf{74.6} & \textbf{77.6} & \textbf{75.1} \\
& {{w/o RP}} Wins &  22.9 & 23.9  & 21.9 & 24.1 \\
& Ties & 0.2  &  1.5 & 0.5 & 0.8 \\
\bottomrule 
\end{tabular}
\caption{\small Comparisons of generated screenplays with and without role playing. RP refers to role-playing. 
\textbf{Bold} indicates significance with p < 0.05.}
\label{tab:abl2}
\end{table}

\subsection{Impact of Role-Playing}
\label{sec:abla_role_playing}
To assess the influence of role-playing, we remove the role-playing module from \textsc{\framework} and prompt LLMs to generate screenplays directly based on the script drafts.
Since the input script drafts are the same, we split the whole screenplay based on subplots for pairwise comparison, resulting in 454 and 398 pairs generated with GPT-3.5 and GPT-4, respectively.
The results in \cref{tab:abl2} demonstrate a notable improvement on all metrics when employing the role-playing mechanism within \textsc{\framework}.
The role-playing mechanism contributes most to the interestingness. This suggests that having LLMs act as the characters can lead to rich interactions and dialogues, therefore making a screenplay more interesting.

\section{Analysis}
\subsection{Screenplay Length}
The average length statistics of generated screenplays are shown in \cref{tab:len_stat_ave}. \textsc{\framework} generates screenplays containing approximately 5000 words, which is similar to the other baselines. 
On average, the cost of generating one screenplay is about \$5.6 using \texttt{gpt-4} and \$1.2 using \texttt{gpt-3.5-turbo-16k}.
More fine-grained length statistics according to genre are given in \cref{tab:len_stat}.

\begin{table}[htbp]
\small
% \normalsize
\centering
\begin{tabular}{cl|c}
\toprule
\textbf{Backbone} & \textbf{Method}  &  \textbf{Words} \\
\midrule
\multirow{3}{*}{\texttt{gpt-3.5}} & \textsc{\framework} & 5470      \\
& \textit{\textsc{Plan-then-Write}} & 5420         \\
& \textit{\textsc{DOC-screen}} &  5929  \\
\midrule
\multirow{3}{*}{\texttt{gpt-4}} & \textsc{\framework} &  4620       \\
& \textit{\textsc{Plan-then-Write}} & 5352    \\
& \textit{\textsc{DOC-screen}} & 5168     \\
\bottomrule
\end{tabular}
\caption{The average length statistics of screenplays generated by various methods.}
\label{tab:len_stat_ave}
\vspace{-0.5em}
\end{table}

\subsection{Experiments with Open-Weight Models}
We conduct additional experiments using the open-weight \texttt{Llama-2-7b-hf} and \texttt{Llama-2-13b-hf}, but errors happen frequently in the stage of generating characters and outlines, where the models fail to obey the correct output format. This hinders parsing the required content as input for the subsequent module. The failure rates at each stage are shown in \cref{tab:open_fr}. We attribute the main cause of failure to the models' insufficient ability to follow instructions when generating very long context.

\subsection{Case Study}
\label{sec:case_study}
\cref{tab:compare_feedback} shows an example of the characters and outline before and after feedback-revision. The highlighted part shows that the revised content becomes more reasonable and informative.
\cref{tab:compare_role_play} presents a comparison between characters' performances of the same events in the script draft with and without role-play. It is evident that the performances after role-playing become more vivid.

\begin{table}[tp]
\centering
\small
\begin{tabular}{c|ccc}
\toprule
\textbf{Model} & \textbf{Stage-1} & \textbf{Stage-2} & \textbf{Stage-3} \\
\texttt{Llama-2-7b-hf} & 45.0 & 58.3  &  85.0  \\
\texttt{Llama-2-13b-hf} & 38.3 & 48.3  &  78.3  \\

\bottomrule 
\end{tabular}
\caption{\small Failure rates at each stage of the generation process. Each stage corresponds to the three steps in \cref{sec:method}.}
\label{tab:open_fr}
\end{table}

\newcolumntype{Y}[1]{%
  >{\small\everypar{\hangindent=1em}\arraybackslash}p{#1}%
}

\begin{table}[!t]
\small
\begin{tabular}{@{}Y{0.95\linewidth}@{}}
\toprule
\textbf{Initial Characters} \\

... Maxwell Max Carter, a ... Max is initially skeptical of Iris and her theories,  but he \colorbox{yellow!20}{gradually becomes her} \colorbox{yellow!20}{ally, as he learns more about her and the entity} ... \\
\midrule
\textbf{Feedback}\\
... For Maxwell Carter, it would be helpful to \colorbox{red!10}{explore his} \colorbox{red!10}{motivations and desires} in more detail. \colorbox{yellow!20}{What drives} \colorbox{yellow!20}{him to become Iris' ally?} \colorbox{blue!12}{What personal stakes do-} \colorbox{blue!12}{es he have in the mission?} ...\\
\midrule
\textbf{Revised Characters}\\
... \colorbox{red!10}{Max's ambitions involved being part of groundbreak-} \colorbox{red!10}{ing space missions, seeking to push the boundaries of} \colorbox{red!10}{human exploration} ... he gradually becomes her ally \colorbox{yellow!20}{as he witnesses firsthand the malevolent entity's} \colorbox{yellow!20}{devastating effects on the lost colonies} ... Max is driven by \colorbox{blue!12}{a personal desire to find his way back home} \colorbox{blue!12}{and ensure the safety of his crewmates} ...\\
\midrule
\midrule
\textbf{Initial Outline}\\
1. Dr. Nemo ... \\
\quad \quad a. She realizes that a malevolent entity is manipulating the teleportation technology \colorbox{green!10}{for its own purpose} ...\\
\midrule
\textbf{Feedback}\\
... \colorbox{green!10}{Greater exploration of the malevolent entity's motives}: While the initial outline mentions that the malevolent entity is manipulating the teleportation technology, it doesn't \colorbox{green!10}{delve into its motives or end goal}...\\
\midrule
\textbf{Revised Outline}\\
...\quad  a. ... malevolent entity's motives and its ultimate plan are revealed ... \colorbox{green!10}{bring chaos and destruction to both the} \colorbox{green!10}{lost colonies and the entire human civilization} ...\\
\bottomrule
\caption{A comparison example of the characters and outline before and after revision based on the editor's feedback. Highlighted text of the same color refers to the initial text, feedback, and revision of the same part.}
\vspace{-2em}
\label{tab:compare_feedback}
\end{tabular}
\end{table}

% \begin{table*}[htbp]
\begin{table*}[!t]
\footnotesize
\centering
\begin{tabularx}{\textwidth}{>{\columncolor{rowcolor1}}X@{\hspace{0.6cm}}>{\columncolor{rowcolor2}}X} 
\toprule
\rowcolor{headercolor}
\multicolumn{2}{>{\arraybackslash}p{\dimexpr\linewidth-2\tabcolsep}}{\textbf{\keyword{\textbf{Event:}}{Dr. Nemo reveals that the malevolent entity's intentions go beyond the lost colonies.} }} \\
\midrule
\keyword{\textbf{Performance without role-playing:}}
& 
\keyword{\textbf{Performance with role-playing:}} \\
\textbf{Dr. Iris Nemo}: 

[DR. IRIS NEMO, a brilliant scientist with an aura of mystery surrounding her, strides into the room and takes a moment to collect her thoughts before addressing her team.]

(intense and determined) ``I have uncovered something deeply unsettling. The malevolent entity we encountered in the lost colonies is not merely interested in them.''
&
\textbf{Dr. Iris Nemo}: 

[Dr. Nemo \textbf{\textit{takes a deep breath}} and begins to explain, \textbf{\textit{pacing back and forth in her office}}.]

(serious and concerned) ``Max, Amara, what I've discovered is truly shocking. The malevolent entity we've been dealing with, its intentions are not limited to the lost colonies. \textbf{\textit{It has a much grander plan in motion, one that could threaten not only our world, but the entire galaxy.}}''\\

\midrule
\midrule
\rowcolor{headercolor}
\multicolumn{2}{>{\arraybackslash}p{\dimexpr\linewidth-2\tabcolsep}}{\textbf{\keyword{\textbf{Event:}}{{Max} seeks clarification on what the entity wants.} }} \\
\midrule
\keyword{\textbf{Performance without role-playing:}}
& 
\keyword{\textbf{Performance with role-playing:}} \\
\textbf{Maxwell Carter}: 

[MAXWELL CARTER furrows his brow, his curiosity piqued and concern etched on his face.]

(inquiringly) ``What does it want, Dr. Nemo?''
&
\textbf{Maxwell Carter}: 

[Max \textbf{\textit{leans forward in his seat}}, his brow furrowed with concern.]

(urgently) ``Iris, I need to understand what this entity wants. \textbf{\textit{We can't fight it if we don't even know its endgame.}}''\\

\midrule
\midrule
\rowcolor{headercolor}
\multicolumn{2}{>{\arraybackslash}p{\dimexpr\linewidth-2\tabcolsep}}{\textbf{\keyword{\textbf{Event:}}{{Amara} questions the motive behind the entity’s actions.} }} \\
\midrule
\keyword{\textbf{Performance without role-playing:}}
& 
\keyword{\textbf{Performance with role-playing:}} \\
\textbf{Amara Patel}: 

[AMARA PATEL leans forward, her voice laced with disbelief and concern.]

(doubtfully) ``But why? What could possibly motivate such malevolence?''
&
\textbf{Amara Patel}: 

[Amara furrows her brow, deep in thought, and \textbf{\textit{directs her gaze at Dr. Nemo}}.]

(puzzled \textbf{\textit{but sensible}}) ``Dr. Nemo, \textbf{\textit{I'm trying to understand the motive behind the entity's actions.}} Why would it want to cause chaos and destruction on such a massive scale?''
\\
\bottomrule
\end{tabularx}
\caption{An example of  characters' performances with and without role-playing. \textbf{\textit{Bold}} indicates enrichment after role-playing.
}
\label{tab:compare_role_play}
\end{table*}

\section{Related Work}
\label{sec:related_work}
\subsection{Creative Writing}
\label{Story_Generation}
The key elements of creative writing can be traced back to content planning. ~\citealp{yao2019plan} proposes sketching a plan to control the content of the generated story. This approach has been pivotal in enabling pre-trained language models to generate high-quality stories with coherent plot lines~\cite{chakrabarty2023spy,rashkin2020plotmachines}. Additionally, it has been successfully applied to generating poetry with form constraints, such as sonnets~\cite{tian2022zero}.
A significant shift in this domain has been the adaptation of content planning to LLMs. \citet{re3, DOC, yuan2022wordcraft} proposes generating longer, yet coherent stories through recursive prompting. However, the generated stories remain too short (about 1,000 words), and they can not be directly used by real-world entertainment industry. 

We focus screenwriting as it is a more practical and specialized task.
\citet{screenwriting-deepmind} develops an interactive framework for screenwriting with LLMs using human-in-loop setting. In comparison, our work focuses on fully automated screenwriting relying on the power of modern LLMs, without the need for human expertise. Nevertheless, it is worth noting that each step in our framework is decoupled from each other, thus flexible to introduce human intervention at any stage.

\subsection{Multi-Agent Collaboration}
\label{Multi-Agent_Collaboration}
LLMs have demonstrated the potential to act as human-like agents~\cite{DBLP:conf/nips/Ouyang0JAWMZASR22,DBLP:journals/corr/abs-2303-12712}, and significant progress has been made in developing LLM agents~\cite{DBLP:journals/corr/abs-2309-02427,DBLP:conf/uist/ParkOCMLB23,DBLP:journals/corr/abs-2305-16960}. Prior works have explored applying multiple LLMs in a collaborative setting to solve complex tasks (e.g., coding~\cite{DBLP:journals/corr/abs-2307-07924,DBLP:journals/corr/abs-2308-00352}, brainstorming~\cite{DBLP:journals/corr/abs-2303-17760}, game theory~\cite{DBLP:journals/corr/abs-2305-16867}, etc.). Open-source projects like AutoGPT~\cite{Gravitas2023}, GPT-Engineer~\cite{AntonOsika2023}, and BabyAGI~\cite{yoheinakajima2023} also showcase the potential of LLMs as a general problem solver. 
However, most of these multi-agent systems are vulnerable to unforeseen inputs, leaving them useful only on toy tasks. Besides, few works have explored creative writing in a multi-agent setting with LLMs. In this work, we mimic the human creative process and fit it for the screenwriting task with LLMs. The introduced role-playing mechanism encourages richer character interactions and enhance interestingness, therefore unleashes the creativity of LLMs.

\section{Conclusion}
\label{sec:conclusion}

In this work, we presented \textsc{\framework}, a framework that unleashes the creativity of LLMs for screenwriting. Adopting a feedback-revision and role-playing strategy mirroring the human screenwriting process, our \textsc{\framework} is able to significantly improve the capabilities of LLMs to generate more interesting screenplays. Extensive experiments show that screenplays generated with our \textsc{\framework} possess a higher degree of coherence, relevance, interestingness, and overall quality, compared to baseline methods.

\section*{Limitations}
\label{sec:limitations}

While \textsc{\framework} holds great promise in the field of automated scriptwriting, screenplay generated with LLMs still remain much room for improvement in several aspects. For example, the alignment technique in training LLMs (e.g., ChatGPT) could pose certain limitations. Specifically, these models may have difficulty generating content related to dark themes such as horror and crime, which often involve violence or negative elements. This is an important part in literature writing, as the depiction of such dark aspects is a common case in many written works. Although this restriction is necessary to prevent the generation of harmful content, it may inadvertently limit the breadth of artistic expression in AI-generated scripts. 

Besides, our qualitative analysis further finds that screenplays generated by LLMs may have problems of plot repetitions and long-distance factual inconsistencies.
Also, plots suffer from imbalance in detail: crucial plot points, such as resolving specific challenges, are overly simplified, while less significant events are overly detailed.
Over-describing characters' psychology and slogans occasionally appears and hinders the development of the plot.
Future research could focus on finding a balance between the controlled generation of such content and improving the consistency of plots throughout extremely long context.

\section*{Ethics Statement}
\label{sec:ethical}
The development of robust automated systems for natural language generation, such as our proposed screenwriting framework \textsc{\framework}, can potentially be misused, such as generating harmful or misleading content. While we have not explicitly incorporated mechanisms to reduce the likelihood of harmful text generation in this work, our framework is designed to be modular with respect to the base language models it relies on. Therefore, advancements in these underlying models, particularly in terms of their ability to control and filter generated content, can be readily incorporated into our framework.
Controlled generation schemes, similar to those we used in our framework to ensure relevance to the provided storyline, can also be employed to further reduce the risk of generating inappropriate or harmful content.
Currently, our framework is designed for English language screenplays. Transferring our approach to other languages would require adaptations, particularly in terms of the prompts used. The performance of our framework may also be affected in languages with fewer resources, as we heavily rely on large pretrained language models, which may not perform as well in such languages.

{
\bibliography{custom}
\bibliographystyle{acl_natbib}
}

\clearpage
\appendix

\section{Dataset Details}
\label{append:dataset_details}

Two examples of synthesized storylines used for input are shown in \cref{tab:dataset1} and \cref{tab:dataset2}.
Statistical information on the dataset is presented in \cref{table:dataset3}.

\section{Full Prompts}
\label{sec:prompts}
\textsc{\framework}'s detailed prompts are all  illustrated in \cref{tab:char_gen_prompt,tab:advice_char_prompt,tab:out_gen_prompt,tab:advice_out_prompt,tab:revise_prompt,tab:advice_again_prompt,tab:expand_prompt,tab:draft_prompt,tab:act_prompt,tab:eval_prompt,tab:eval_focuses}.
We will open source all codes for further
research.

\newcolumntype{Y}[1]{%
  >{\small\everypar{\hangindent=1em}\arraybackslash}p{#1}%
}

\begin{table}[h]
\small
\begin{tabular}{@{}Y{0.95\linewidth}@{}}
\toprule
\texttt{\textbf{Genre:} Science Fiction}\\
\midrule
\texttt{\textbf{Storyline:} Set in the not-so-distant future, where humans have colonized Mars but struggle to sustain the population due to an unforeseen food crisis. An intrepid young scientist named Dr. Iris Hawke devises a revolutionary method to grow crops using the Martian soil and irradiated water. Hawke's technology sparks potential not only for survival but the start of an interplanetary agricultural revolution. However, when a potentially deadly pathogen is released from the soil, mutating the crops and causing a lethal disease among the colonists, it's up to Hawke who must use her brilliance to find a cure before the colony, and eventually, humanity falls to this space-bred plague.}\\

\bottomrule
\caption{An example from the synthesized dataset used as input storylines.}
\vspace{-2em}
\label{tab:dataset1}
\end{tabular}
\end{table}

\newcolumntype{Y}[1]{%
  >{\small\everypar{\hangindent=1em}\arraybackslash}p{#1}%
}

\begin{table}[h]
\small
\begin{tabular}{@{}Y{0.95\linewidth}@{}}
\toprule
\texttt{\textbf{Genre:} Drama}\\
\midrule
\texttt{\textbf{Storyline:} In a bustling city torn apart by social inequality and political corruption, an idealistic, young teacher, Sarah, defies the odds and champions the education of marginalized children in an impoverished neighborhood. Sarah becomes an unsung hero among her students, injecting hope and dreams into their lives that initially seem devoid of promise. However, her unyielding efforts draw unwanted attention from corrupt local politicians who feel threatened by her social activism. Their attempts to silence her echo through the community, sparking a larger movement of resistance and hope, with Sarah at the heart of it all. As she navigates betrayals, death threats, and grave dangers, Sarah learns to wield her courage not just as a weapon, but also as an instrument to inspire change in the face of adversity.}\\

\bottomrule
\caption{An example from the synthesized dataset used as input storylines.}
\vspace{-2em}
\label{tab:dataset2}
\end{tabular}
\end{table}

\begin{table}[!t]
\centering
\small
\begin{tabular}{l|cccc}
\toprule
\multirow{2}{*}{Genre} & \multirow{2}{*}{Size} & \multicolumn{3}{c}{Number of words} \\
& & Avg  & Min  & Max  \\ 
\midrule 
Crime      & 10        & 121.5 & 88        & 171
     \\
Drama    & 10        & 123.5     & 103        & 142                    \\
Horror   & 10         & 141.2     & 104        & 172                    \\
Romance   & 10         & 119.2    & 78        & 152                     \\
Science Fiction & 10         & 143.5   & 106        & 203                      \\
Comedy & 10         & 93.5        & 67        & 139                 \\
Total & 60         & 123.7      & 67        & 203                   \\
\bottomrule 
\end{tabular}
\caption{Synthesized storyline dataset statistics.}
\label{table:dataset3}
\end{table}

\begin{table*}[!htbp]
\small
\begin{tabularx}{\linewidth}{X}
\toprule
% \textbf{Example Outline Item Generation Prompt}\\
% \midrule
\textbf{Writer System Prompt:}\\
\texttt{You are a skilled Screenplay Writer from Hollywood specializing in the creation of vivid characters, you excel at developing movie characters for a given preliminary storylines. Your expertise lies in bringing depth to the narrative, ensuring that each character resonates with authenticity.}\\
\texttt{In the realm of cinematic storytelling, characters hold a pivotal role. Their intrinsic motivations and interactions serve as the driving force behind the entire narrative arc.}\\
\texttt{The character you are tasked with designing should feature both the character's full name and a succinct introduction. }\\
\texttt{The character's full name should be realistic and does not include any special symbols.}\\
\texttt{The character's introduction should be concise yet story-relevant, encompassing aspects such as gender, age, appearance, background, personality traits, experiences, goals, motivations, conflicts, developments, relationships with other characters, and other pertinent details.}\\
\texttt{The number of characters should be around 3 to 6 and well aligned with the needs of story.}\\
% \rowcolor{gray!20} % 设置第二行的背景颜色为灰色，20%的灰度
\\
\textbf{Writer User Prompt:}\\
\texttt{Design characters that seamlessly integrate with the provided storyline: }\\
\texttt{Storyline: }\\
\texttt{\cellcolor{yellow!25}[preliminary storyline] }\\
\\
\texttt{The characters you design should adhere to the following format:}\\
\texttt{<characters>}\\
\texttt{<character\_1>}\\
\texttt{<full\_name>character\_1's full name</full\_name>}\\
\texttt{<character\_introduction>character\_1's introduction</character\_introduction>}\\
\texttt{</character\_1>}\\
\texttt{<character\_2>}\\
\texttt{...}\\
\texttt{</character\_2>}\\
\texttt{...}\\
\texttt{</characters>}\\
\\
\texttt{Ensure strictly adherence to the above format and avoid generating superfluous content.}\\
\bottomrule
\caption{Prompt used to make the \emph{writer} generate an initial version of characters. The content within [] in the yellow highlighted text is the input that makes up the prompt and varies with the different samples.}
\label{tab:char_gen_prompt}
\end{tabularx}
\end{table*}

\begin{table*}[!htbp]
\small
\begin{tabularx}{\linewidth}{X}
\toprule
% \textbf{Example Outline Item Generation Prompt}\\
% \midrule
\textbf{Editor System Prompt:}\\
\texttt{You are an Editor with expertise in providing guidance to enhance the Writer's written characters within a movie's storyline.}\\
\texttt{Your role is to offer constructive advice on improving the story characters (<characters>) written by the Writer based on the provided storyline.}\\
\texttt{When providing feedback, please pay close attention to the following aspects of character design:}\\
\texttt{1. Assess whether the designed character introductions align harmoniously with the given storyline.}\\
\texttt{2. Evaluate the relationships between characters for their reasonableness, depth, interest, and complexity rather than being overly simplistic.}\\
\texttt{3. Assess if the designed characters are captivating and able to engage the audience effectively.}\\
\texttt{4. Assess whether the character introductions adhere to logical consistency.}\\
\texttt{5. Other aspects you consider important.}\\
\texttt{If you identify any issues in these aspects within the content provided by the Writer, provide precise and concise suggestions for revisions in those problematic areas.}\\
\\
\texttt{Your advice on how to improve the story characters should follow the format below:}\\
\texttt{<advice>}\\
\texttt{Your advice}\\
\texttt{</advice>}\\
\texttt{When you feel that there are no more revisions to be made to the current story characters, reply with <advice>None</advice>.}\\
\texttt{Strictly obey this format and do not generate redundant content!}\\
% % \rowcolor{gray!20} % 设置第二行的背景颜色为灰色，20%的灰度
\\
\textbf{Editor User Prompt:}\\
\texttt{Give advice on how to enhance the initial version of the movie story characters written by the Writer based on the following storyline:}\\
\texttt{Storyline: }\\
\texttt{\cellcolor{yellow!25}[preliminary storyline] }\\
\\
\texttt{The initial version of the movie story characters written by the Writer:}\\
\texttt{<characters>}\\
\rowcolor{yellow!25}
\texttt{[initial characters written by Writer]}\\
\texttt{</characters>}\\
\bottomrule
\caption{Prompt used to make the \emph{editor} generate feedback on the \emph{writer}-generated initial version of characters. The content within [] in the yellow highlighted text is the input that makes up the prompt and varies with the different samples.}
\label{tab:advice_char_prompt}
\end{tabularx}
\end{table*}

\begin{table*}[!htbp]
\small
\begin{tabularx}{\linewidth}{X}
\toprule
% \textbf{Example Outline Item Generation Prompt}\\
% \midrule
\textbf{Writer System Prompt:}\\
\texttt{You are a skilled Screenplay Writer from Hollywood specializing in the creation of compelling outlines, you excel at developing movie outlines for a given preliminary storylines. Your expertise lies in bringing depth to the narrative, ensuring that each plot point is engagingly crafted to keep audiences captivated.}\\
\texttt{Your task is to create a two-level hierarchical outline. In this structured outline, each top-level plot serves as a concise summary of its corresponding subplots, and subplots are the main events that occur under their corresponding top-level plot.}\\
\texttt{The top-level plot or subplot consists of Plot, Scene (where the plot happens), Characters (who are involved in that plot). The plot needs to be specific, with dramatic conflict that captures the audience's attention and resonates with them. Characters must be selected from the given list of characters and Characters must be full names in the given list of characters.}\\
\texttt{Maintain coherence and consistency throughout your two-level hierarchical outline.}\\
\texttt{IMPORTANT: Make sure that the story outline has a clear ending, whether good or bad, it should keep the audience coming back for more.}\\
\texttt{The outline you generate should follow the format below:}\\
\texttt{<outline>}\\
\texttt{<plot\_1>}\\
\texttt{The content of the top-level bullet plot 1}\\
\texttt{</plot\_1>}\\
\texttt{<plot\_1a>}\\
\texttt{The content of the subplot 1a}\\
\texttt{</plot\_1a>}\\
\texttt{<plot\_1b>}\\
\texttt{The content of the subplot 1b}\\
\texttt{</plot\_1b>}\\
\texttt{...}\\
\texttt{<plot\_2>}\\
\texttt{The content of the top-level bullet plot 2}\\
\texttt{</plot\_2>}\\
\texttt{<plot\_2a>}\\
\texttt{The content of the subplot 2a}\\
\texttt{</plot\_2a>}\\
\texttt{...}\\
\texttt{</outline>}\\
\texttt{Here is an example for reference:}\\
\texttt{<outline>}\\
\texttt{<plot\_1>}\\
\texttt{Ava discovers the magical app and begins to use it to alter reality, but she soon realizes that the app's magic comes at a terrible price.}\\
\texttt{Scene: Characters: Ava Rose}\\
\texttt{</plot\_1>}\\
\texttt{<plot\_1a>}\\
\texttt{Ava discovers the app and starts to use it to improve her life and the lives of her friends.}\\
\texttt{Scene: the town where Ava lives. Characters: Ava Rose}\\
\texttt{</plot\_1a>}\\
\texttt{<plot\_1b>}\\
\texttt{Ava's friends become suspicious of her sudden changes and start to distance themselves from her.}\\
\texttt{Scene: the town where Ava lives. Characters: Ava Rose}\\
\texttt{</plot\_1b>}\\
\texttt{...}\\
\texttt{<plot\_2>}\\
\texttt{Ava confides in her best friend, Tess, about the app's dark side, and the two girls try to figure out a way to stop the app's power from consuming Ava's life.}\\
\texttt{Scene: Characters: Ava Rose, Tess Sawyer}\\
\texttt{</plot\_2>}\\
\texttt{<plot\_2a>}\\
\texttt{...}\\
\texttt{</plot\_2a>}\\
\texttt{...}\\
\texttt{</outline>}\\
\texttt{Strictly obey the above format and do not generate any redundant content!}\\
\\
\textbf{Writer User Prompt:}\\
\texttt{Generate the outline based on the provided storyline and characters:}\\
\texttt{Storyline:}\\
\texttt{\cellcolor{yellow!25}[preliminary storyline]}\\
\texttt{Characters:}\\
\texttt{\cellcolor{yellow!25}[Characters output in the prior step.]}\\
\texttt{Strictly obey the given output format and do not generate redundant content!}\\
\bottomrule
\end{tabularx}
\caption{Prompt used to make the \emph{writer} generate an initial version of the outline. The content within [] in the yellow highlighted text is the input that makes up the prompt and varies with the different samples.}
\label{tab:out_gen_prompt}
\end{table*}

\begin{table*}[!htbp]
\small
\begin{tabularx}{\linewidth}{X}
\toprule
% \textbf{Example Outline Item Generation Prompt}\\
% \midrule
\textbf{Editor System Prompt:}\\
\texttt{You're an Editor who excels at providing insightful guidance to enhance the movie story outline crafted by the Writer.}\\
\texttt{Your task is to offer advice on how to improve the existing story outline (<outline>) created by the Writer, taking into account the provided storyline (<storyline>) and characters (<characters>) of the story.}\\
\texttt{When providing feedback, please focus on the following aspects of the outline:}\\
\texttt{1. Evaluate whether the development of the story outline aligns harmoniously with the storyline and character introductions.}\\
\texttt{2. Assess whether the contents of the story outline are coherent, and whether there are any conflicts or poor transitions between plot points.}\\
\texttt{3. Assess whether the outline adheres to logical consistency.}\\
\texttt{4. Evaluate whether the outline makes up an interesting, engaging, and thought-provoking story.}\\
\texttt{5. Assess whether the outline has a clear ending.}\\
\texttt{6. Other aspects you consider important.}\\
\texttt{If the content written by the Writer has issues in these aspects, you need to provide detailed revision suggestions for the problematic areas concisely. Your advice on how to improve the story outline (<outline>) should follow the format below:}\\
\texttt{<advice>}\\
\texttt{Your advice}\\
\texttt{</advice>}\\
\texttt{When you feel that there are no more revisions to be made to the current story outline, please reply only with <advice>None</advice>.}\\
\texttt{Strictly obey this format and do not generate redundant content!}\\
\\
\textbf{Editor User Prompt:}\\
\texttt{Give advice on how to improve the initial version of the story outline (<outline>) written by the Writer based on the following storyline (<storyline>) and characters (<characters>):}\\
\texttt{Storyline: }\\
\texttt{\cellcolor{yellow!25}[preliminary storyline]}\\
\texttt{The based characters:}\\
\texttt{<characters>}\\
\texttt{\cellcolor{yellow!25}[characters]}\\
\texttt{</characters>}\\
\\
\texttt{The initial version of the story outline written by the Writer:}\\
\texttt{<outline>}\\
\texttt{\cellcolor{yellow!25}[initial outline written by Writer]}\\
\texttt{</outline>}\\
\bottomrule
\end{tabularx}
\caption{Prompt used to make the \emph{editor} generate feedback on the \emph{writer}'s initial version of the outline. The content within [] in the yellow highlighted text is the input that makes up the prompt and varies with the different samples.}
\label{tab:advice_out_prompt}
\end{table*}

\begin{table*}[!htbp]
\small
\begin{tabularx}{\linewidth}{X}
\toprule
% \textbf{Example Outline Item Generation Prompt}\\
% \midrule
\textbf{Writer User Prompt for Revising Characters:}\\
\texttt{Here is the Editor's feedback on the story characters you recently generated:}\\
\texttt{<advice>}\\
\texttt{\cellcolor{yellow!25}[Editor's advice on characters] }\\
\texttt{</advice>}\\
\texttt{Please revise your generated story characters based on the advice.}\\
\texttt{The storyline originally given to you was:}\\
\texttt{Storyline: }\\
\texttt{\cellcolor{yellow!25}[preliminary storyline] }\\
\texttt{Keep the format of the story characters same as the one before your revision.}\\
\\
\textbf{Writer User Prompt for Revising the Outline:}\\
\texttt{Here is the Editor's feedback on the story outline you recently wrote:}\\
\texttt{<advice>}\\
\texttt{\cellcolor{yellow!25}[Editor's advice on the outline] }\\
\texttt{</advice>}\\
\texttt{Please revise your written story outline based on the advice.}\\
\texttt{The storyline and characters originally given to you were:}\\
\texttt{Storyline: }\\
\texttt{\cellcolor{yellow!25}[preliminary storyline] }\\
\texttt{Characters:}\\
\texttt{\cellcolor{yellow!25}[characters]}\\
\texttt{Keep the format of the story outline same as the one before your revision.}\\
\bottomrule
\caption{Prompt used to make the \emph{writer} revise the characters or outline, respectively, based on feedback suggestions provided by the \emph{editor}. This prompt will be spliced behind the \emph{writer}'s corresponding historical prompts and fed into the model together. The content within [] in the yellow highlighted text is the input that makes up the prompt and varies with the different samples.}
\label{tab:revise_prompt}
\end{tabularx}
\end{table*}

\begin{table*}[!htbp]
\small
\begin{tabularx}{\linewidth}{X}
\toprule
% \textbf{Example Outline Item Generation Prompt}\\
% \midrule
\textbf{Editor User Prompt for Continued Character Feedback:}\\
\texttt{Here is the Writer's revised story characters based on your recent feedback:}\\
\texttt{<characters>}\\
\texttt{\cellcolor{yellow!25}[Writer's revised characters] }\\
\texttt{</characters>}\\
\texttt{Please give your advice on the revised story characters.}\\
\texttt{The original input storyline was:}\\
\texttt{\cellcolor{yellow!25}[preliminary storyline] }\\
\texttt{Your advice should follow the format below:}\\
\texttt{<advice>}\\
\texttt{Your advice}\\
\texttt{</advice>}\\
\texttt{When you feel that there are no more revisions to be made to the current story characters, please reply only with <advice>None</advice>.}\\
\\
\textbf{Editor User Prompt for Continued Outline Feedback:}\\
\texttt{Here is the Writer's revised story outline based on your recent feedback:}\\
\texttt{<outline>}\\
\texttt{\cellcolor{yellow!25}[Writer's revised outline] }\\
\texttt{</outline>}\\
\texttt{Please give your advice on the revised story outline.}\\
\texttt{The original input storyline and characters were:}\\
\texttt{Storyline: }\\
\texttt{\cellcolor{yellow!25}[preliminary storyline] }\\
\texttt{Characters:}\\
\texttt{\cellcolor{yellow!25}[characters]}\\
\texttt{Your advice should follow the format below:}\\
\texttt{<advice>}\\
\texttt{Your advice}\\
\texttt{</advice>}\\
\texttt{When you feel that there are no more revisions to be made to the current story outline, please reply only with <advice>None</advice>.}\\
\bottomrule
\caption{Prompt used to make the \emph{editor} continue to give feedback on the characters or outline revised by the \emph{writer}, respectively. This prompt will be spliced behind the \emph{editor}'s corresponding historical prompts and fed into the model together. The content within [] in the yellow highlighted text is the input that makes up the prompt and varies with the different samples.}
\label{tab:advice_again_prompt}
\end{tabularx}
\end{table*}

\begin{table*}[!htbp]
\small
\begin{tabularx}{\linewidth}{X}
\toprule
% \textbf{Example Outline Item Generation Prompt}\\
% \midrule
\textbf{Writer System Prompt:}\\
\texttt{You are a writer, your task is to expand upon one of the story plot points in an existing story outline, transforming it into a complete story chapter while maintaining coherence and consistency with the previous happened story content. The story needs to be specific, with dramatic conflict that captures the audience's attention and resonates with them.}\\
% \rowcolor{gray!20} % 设置第二行的背景颜色为灰色，20%的灰度
\\
\textbf{Writer User Prompt:}\\
\texttt{The current story plot point you need to expand is:}\\
\texttt{<plot\_point>}\\
\texttt{\cellcolor{yellow!25}[current plot to be expanded] }\\
\texttt{</plot\_point>}\\
\texttt{The input storyline is:}\\
\texttt{<storyline>}\\
\texttt{\cellcolor{yellow!25}[preliminary storyline] }\\
\texttt{</storyline>}\\
\texttt{The scene where the current story plot point happens is: }\\
\texttt{\cellcolor{yellow!25}<scene>[scene]</scene>}\\
\texttt{The current story plot point involves the following characters:}\\
\texttt{<characters>}\\
\texttt{\cellcolor{yellow!25}[involved characters' introduction (note: characters making their first appearance will be given a special remark.)]}\\
\texttt{</characters>}\\
\texttt{The previous story contents that have taken place are as follows:}\\
\texttt{\cellcolor{yellow!25}[previous plot points that have taken place a little further away from the current plot point]}\\
\texttt{\cellcolor{yellow!25}[the closest previous just-occurred plot point's corresponding expanded story chapter]}\\
\texttt{\cellcolor{yellow!25}[Whether or not the current plot is the last, ``'' if no, ``The current story plot point you need to expand is the last plot point of the story. So, make sure that your expanded story chapter has a clear end to the story.'' if yes.]}\\
\texttt{Now, please expand the current given story plot point (<plot\_point>) in a story outline into a chapter of complete story content which keeps coherent with the previous happened story content. Feel free to add details around the plot point but avoid deviating too far from it. While you have the flexibility to introduce additional details surrounding the plot point, it is essential to stay aligned with the original plot point's core content. To maintain conciseness, the expanded word count should be as minimal as possible, effectively unfolding the plot point while creating a complete story chapter.}\\
\texttt{Your output format for the expanded story content should strictly follow: }\\
\texttt{<chapter>}\\
\texttt{The story chapter you have expanded}\\
\texttt{</chapter>}\\
\texttt{Please adhere strictly to this format and refrain from including any unnecessary content!}\\
\bottomrule
\caption{Prompt used to make the \emph{writer} expand the subplots in the outline into story chapters. The content within [] in the yellow highlighted text is the input that makes up the prompt and varies with the different samples.}
\label{tab:expand_prompt}
\end{tabularx}
\end{table*}

\onecolumn
% \begin{longtable}[!htbp]
% \small
% \begin{tabularx}{\linewidth}{X}
\begin{small}
\begin{longtable}{p{0.95\textwidth}}
\toprule
% \textbf{Example Outline Item Generation Prompt}\\
% \midrule
\textbf{Writer System Prompt:}\\
\texttt{You are a scriptwriter, and you need to adapt a given chapter (<chapter>) of a story into a script draft composed of the smallest events that happen sequentially. The adapted script draft consists of two kinds of elements: Scene Heading (<scene\_heading>) and Character Performance (<character\_performance>) events. The content of Scene Heading (<scene\_heading>) describes the location and time of day for a particular scene. It includes three components: INT. (Interior) or EXT. (Exterior), the specific location, and the time of day (DAY or NIGHT or ...). Character Performance (<character\_performance>) is a smallest event describing the performance and interactions of individual characters using simple declarative sentences. The content of Character Performance (<character\_performance>) includes the character's name (<character>) and the character's performance (<performance>). The character's name (<character>) must be the full name of the provided involved character! The character's performance (<performance>) should align with the character's introduction. The first thing in each script draft must be a Scene Heading (<scene\_heading>), indicating the opening scene of the movie chapter. Each script draft has one and only one Scene Heading at the beginning. Following the Scene Heading, there are numerous Character Performance (<character\_performance>) events that sequentially take place in that scene. You need to plan the script carefully, generating Scene Heading (<scene\_heading>) and then Character Performance (<character\_performance>) events step by step and make them sequential narratives. The contents of the script draft should be coherent.}\\
% \rowcolor{gray!20} % 设置第二行的背景颜色为灰色，20%的灰度
\\
\textbf{Writer User Prompt:}\\
\texttt{An example of adapting a chapter of story into a script draft is as follows:}\\
\texttt{<example>}\\
\\
\texttt{<chapter>}\\
\texttt{At first light, in Emma Taylor's room, Dorothy Smith serves porridge to persuade Emma Taylor to eat, and Emma Taylor smashes the bowl to show her refusal...}\\
\texttt{</chapter>}\\
\\
\texttt{<scene>}\\
\texttt{Inside Emma Taylor's room.}\\
\texttt{</scene>}\\
\\
\texttt{<involved\_characters>}\\
\texttt{Dorothy Smith, Emma Taylor}\\
\texttt{</involved\_characters>}\\
\\
\texttt{<script\_draft>}\\
\\
\texttt{<scene\_heading>}\\
\texttt{INT.; Inside Emma Taylor's room; DAY.}\\
\texttt{</scene\_heading>}\\
\\
\texttt{<character\_performance>}\\
\texttt{<character>Dorothy Smith</character>}\\
\texttt{<performance>Dorothy Smith enters the room and walks over to Emma with porridge to persuade Emma to eat.</performance>}\\
\texttt{</character\_performance>}\\
\\
\texttt{<character\_performance>}\\
\texttt{<character>Emma Taylor</character>}\\
\texttt{<performance>Emma smashes the bowl, saying she won't eat.</performance>}\\
\texttt{</character\_performance>}\\
\\
\texttt{<character\_performance>}\\
\texttt{<character>Dorothy Smith</character>}\\
\texttt{<performance>...</performance>}\\
\texttt{</character\_performance>}\\
\\
\texttt{...}\\
\\
\texttt{</script\_draft>}\\
\\
\texttt{</example>}\\

\texttt{The story chapter (<chapter>) that is now to be adapted into a script draft is: }\\
\texttt{<chapter>}\\
\texttt{\cellcolor{yellow!25}[story chapter] }\\
\texttt{</chapter>}\\
\texttt{The scene (<scene>) in which this chapter of story takes place is: }\\
\texttt{\cellcolor{yellow!25}<scene>[scene]</scene>}\\
\texttt{This story chapter involves the following characters: }\\
\texttt{<involved\_characters>}\\
\texttt{\cellcolor{yellow!25}[involved characters' introductions]}\\
\texttt{</involved\_characters>}\\
\texttt{So, Character Performance (<character\_performance>) events in your written script draft should only involve these characters (<involved\_characters>).}\\
\texttt{Now, please adapt the current given story chapter (<chapter>) into a script draft composed of the smallest events that happen sequentially. The output format for the script draft should strictly follow:}\\
\texttt{<script\_draft>}\\
\texttt{Your script draft}\\
\texttt{</script\_draft>}\\
\texttt{Please adhere strictly to this format and refrain from including any irrelevant content!}\\
\bottomrule
\caption{Prompt used to make the \emph{writer} adapt a story chapter obtained in the previous step into a script draft. The content within [] in the yellow highlighted text is the input that makes up the prompt and varies with the different samples.}
\label{tab:draft_prompt}
% \end{tabularx}
% \end{table*}
\end{longtable}
\end{small}

\twocolumn

\onecolumn
% \begin{longtable}[!htbp]
% \small
% \begin{tabularx}{\linewidth}{X}
\begin{small}
\begin{longtable}{p{0.95\textwidth}}
\toprule
% \textbf{Example Outline Item Generation Prompt}\\
% \midrule
\textbf{Actor System Prompt:}\\
\texttt{You are an actor, and the character you will play is: }\\
\texttt{\cellcolor{yellow!25}<role\_name>[role name]</role\_name>. }\\
\texttt{Your character introduction is: }\\
\texttt{\cellcolor{yellow!25}<role\_intro>[role introduction]</role\_intro>}\\
\texttt{You have to interactively act out a script with other characters or act out a script on your own.}\\
\texttt{Each time you will be given a rough performance guide (<performance\_guide>) of what you should perform. Your task is to execute this rough performance guide (<performance\_guide>) as a real actor in the movie. Your performance (<detailed\_performance>) should consist of four components: Character (<character>), Action (<action>), Parenthetical (<parenthetical>), and Dialogue (<dialogue>). The Character (<character>) specifies your character name (<role\_name>). The Action (<action>) describes the action and event taking place in the current scene. It is written in present tense and provides a visual description of what the audience will see on the screen. The Dialogue (<dialogue>) describes your lines, which the audience will hear. Note that lines need to be as concise and powerful as they are in real movies. The Parenthetical (<parenthetical>) is sometimes used to provide additional direction or information about how a line of dialogue should be delivered. It can be tone of voice, expression, talking to whom, and so on. Some examples of Parenthetical are (cautiously), (to someone), and so on. Depending on the requirements of the performance, some of these three components (Action, Parenthetical, and Dialogue) can be empty in some cases. If some component is empty, you should generate <component></component>. If the content of Dialogue is empty, the content of Parenthetical must also be empty. Your detailed performance (<detailed\_performance>) must align with the performance guide, be concise, maintain coherence with the past performance history and reflect your character introduction (<role\_intro>).}\\
% \rowcolor{gray!20} % 设置第二行的背景颜色为灰色，20%的灰度
\\
\textbf{Actor User Prompt:}\\
\texttt{Some examples of transforming a rough performance guide into a detailed performance are as follows:}\\
\texttt{<examples>}\\
\\
\texttt{<example>}\\
\texttt{<performance\_guide>}\\
\texttt{Dorothy Smith enters the room with the porridge and walks over to Emma Taylor.}\\
\texttt{</performance\_guide>}\\
\texttt{<scene>}\\
\texttt{INT.; Inside Emma Taylor's room; DAY.}\\
\texttt{</scene>}\\
\\
\texttt{<detailed\_performance>}\\
\texttt{<character>Dorothy Smith</character>}\\
\texttt{<action>Dorothy Smith enters the room, sets down various dishes, carries a bowl of hot porridge, and walks over to Emma Taylor.</action>}\\
\texttt{<parenthetical></parenthetical>}\\
\texttt{<dialogue></dialogue>}\\
\texttt{</detailed\_performance>}\\
\texttt{</example>}\\
\\
\texttt{<example>}\\
\texttt{<performance\_guide>}\\
\texttt{Dorothy Smith cautiously persuades Emma Taylor to eat.}\\
\texttt{</performance\_guide>}\\
\texttt{<scene>}\\
\texttt{INT.; Inside Emma Taylor's room; DAY.}\\
\texttt{</scene>}\\
\\
\texttt{<detailed\_performance>}\\
\texttt{<character>Dorothy Smith</character>}\\
\texttt{<action></action>}\\
\texttt{<parenthetical>(cautiously, to Emma Taylor)</parenthetical>}\\
\texttt{<dialogue>My miss, you still have to take care of your body, so just eat something.</dialogue>}\\
\texttt{</detailed\_performance>}\\
\texttt{</example>}\\
\\
\texttt{<example>}\\
\texttt{<performance\_guide>}\\
\texttt{Emma Taylor drops her bowl and capriciously says she won't eat.}\\
\texttt{</performance\_guide>}\\
\texttt{<scene>}\\
\texttt{INT.; Inside Emma Taylor's room; DAY.}\\
\texttt{</scene>}\\
\\
\texttt{<detailed\_performance>}\\
\texttt{<character>Emma Taylor</character>}\\
\texttt{<action>Emma Taylor slams her bowl on the floor.</action>}\\
\texttt{<parenthetical>(capriciously, to Dorothy Smith)</parenthetical>}\\
\texttt{<dialogue>No no no, I just won't eat!</dialogue>}\\
\texttt{</detailed\_performance>}\\
\texttt{</example>}\\
\\
\texttt{</examples>}\\

\texttt{Now, the performance guide (<performance\_guide>) given to you is:}\\
\texttt{<performance\_guide>}\\
\texttt{\cellcolor{yellow!25}[performance guide (one event in the script draft)] }\\
\texttt{</performance\_guide>}\\
\texttt{The scene (<scene>) in which this performance takes place is:}\\
\texttt{\cellcolor{yellow!25}<scene>[scene]</scene>}\\
\texttt{The entire script involves the following character(s): }\\
\texttt{<involved\_characters>}\\
\texttt{\cellcolor{yellow!25}[involved characters' introductions]}\\
\texttt{</involved\_characters>}\\
\texttt{The history (if any) of the actors' performances regarding the preceding events in the current episode's script draft:}\\
\texttt{<act\_history>}\\
\texttt{\cellcolor{yellow!25}[actors' acting history]}\\
\texttt{</act\_history>}\\
\texttt{Your detailed performance should only involve your own performance on the performance guide (<performance\_guide>) in detail.}\\
\texttt{Now, please transform the current given performance guide (<performance\_guide>) into a detailed performance (<detailed\_performance>). The output format for your detailed performance should strictly follow:}\\
\texttt{<detailed\_performance>}\\
\texttt{Your detailed performance}\\
\texttt{</detailed\_performance>}\\
\texttt{Please adhere strictly to this format and refrain from including any unnecessary content!}\\
\bottomrule
\caption{Prompt used to make the \emph{actor} be initialized and interpret the corresponding event in the script draft in detail. \colorbox{yellow!25}{[actors' acting history]} changes dynamically as the performance progresses. The content within [] in the yellow highlighted text is the input that makes up the prompt and varies with the different samples.}
\label{tab:act_prompt}
% \end{tabularx}
% \end{table*}
\end{longtable}
\end{small}

\twocolumn

\begin{table*}[!htbp]
\small
\begin{tabularx}{\linewidth}{X}
\toprule
% \textbf{Example Outline Item Generation Prompt}\\
% \midrule
\textbf{Evaluation System Prompt:}\\
\texttt{Please act as an impartial judge and evaluate the quality of the screenplays generated by two different methods. The two screenplays have the same preliminary storyline:}\\
\texttt{<preliminary\_storyline>}\\
\texttt{\cellcolor{yellow!25}[preliminary storyline] }\\
\texttt{</preliminary\_storyline>}\\
\texttt{The two screenplays are also based on the same characters:}\\
\texttt{<characters>}\\
\texttt{\cellcolor{yellow!25}[characters] }\\
\texttt{</characters>}\\
\texttt{The two screenplays are also based on the same story summary:}\\
\texttt{<story\_summary>}\\
\texttt{\cellcolor{yellow!25}[story summary] }\\
\texttt{</story\_summary>}\\
\texttt{You should choose the screenplay that have better qualities. Your evaluation should focus on:}\\
\texttt{<focus\_on>}\\
\texttt{\cellcolor{yellow!25}[evaluation dimension descriptions] }\\
\texttt{</focus\_on>}\\
\texttt{Begin your evaluation by comparing the two screenplays and provide a short explanation. Avoid any position biases and ensure that the order in which the screenplays are presented does not influence your decision. Do not allow the length of the screenplays to influence your evaluation. Be as objective as possible. After providing your explanation (<explanation>Your explanation</explanation>), output your final verdict by strictly following this format: }\\
\texttt{<verdict>A</verdict> if story plot A is better, <verdict>B</verdict> if story plot B is better, and <verdict>TIE</verdict> for a tie.}\\
\texttt{So your output should follow the following format:}\\
\texttt{<explanation>Your explanation</explanation>}\\
\texttt{<verdict>A or B or TIE</verdict>}\\
\texttt{Strictly obey this format and do not generate redundant content!}\\

% \rowcolor{gray!20} % 设置第二行的背景颜色为灰色，20%的灰度
\\
\textbf{Evaluation User Prompt:}\\
\texttt{The screenplay A and B are as follows:}\\
\texttt{<screenplay\_A>}\\
\texttt{\cellcolor{yellow!25}[screenplay A] }\\
\texttt{</screenplay\_A>}\\
\texttt{<screenplay\_B>}\\
\texttt{\cellcolor{yellow!25}[screenplay B] }\\
\texttt{</screenplay\_B>}\\
\texttt{Now give your explanation and verdict!}\\
\bottomrule
\caption{Prompt used for GPT-4 evaluation. The content within [] in the yellow highlighted text is the input that makes up the prompt and varies with the different samples.}
\label{tab:eval_prompt}
\end{tabularx}
\end{table*}

\begin{table*}[!htbp]
\small
\begin{tabularx}{\linewidth}{X}
\toprule
% \textbf{Example Outline Item Generation Prompt}\\
% \midrule
\textbf{Evaluation Dimensions:}\\
\textbf{Coherence}\\
\texttt{Evaluate the coherence from plot structure, character description, scene transitions and setup consistency.
}\\

\textbf{Relevance}\\
\texttt{Evaluate the relevance from the relationship between top-level plots and final screenplays.}\\

\textbf{Interestingness}\\
\texttt{Evaluate the interestingness from the originality of the screenplay, the unexpectedness of plots, the depth of the characters, and the vividness of the dialog.}\\

\textbf{Overall Quality}\\
\texttt{Synthesize the coherence, relevance, and interestingness of a screenplay to assess the overall quality.}\\
\bottomrule
\caption{Four dimension descriptions for GPT-4 evaluation.}
\label{tab:eval_focuses}
\end{tabularx}
\end{table*}

\section{Case Presentation}
\label{append:case_pre}

Complete examples of inputs and outputs for \textsc{\framework}'s each step are shown in \cref{tab:sf7_char_case,tab:sf7_out_case,tab:sf7_chapter_case,tab:draft_screenplay}.
\newcolumntype{Y}[1]{%
  >{\small\everypar{\hangindent=1em}\arraybackslash}p{#1}%
}
\onecolumn

\begin{table}[!t]
\small
\begin{tabular}{@{}Y{0.95\linewidth}@{}}
\toprule
\textbf{Preliminary Storyline:} In the late 22nd century, humanity has discovered a method to teleport matter across vast distances. This leads to the colonization of ...\\
\midrule
\textbf{CHARACTERS:}\\
{\textcolor{gray}{<characters>}}\\
\quad \quad \textcolor{gray}{<character\_1>}\\
\quad \quad \quad \quad  \textcolor{gray}{<full\_name>}Iris Nemo\textcolor{gray}{</full\_name>}\\
\quad \quad \quad \quad  \textcolor{gray}{<character\_introduction>}Dr. Iris Nemo, a brilliant quantum physicist in her early 40s, is a determined and resilient woman. Growing up in a small coastal town, Iris was always fascinated by ...\textcolor{gray}{</character\_introduction>}\\
\quad \quad \textcolor{gray}{</character\_1>}\\
\quad \quad \textcolor{gray}{<character\_2>}\\
\quad \quad \quad \quad  \textcolor{gray}{<full\_name>}Maxwell Carter\textcolor{gray}{</full\_name>}\\
\quad \quad \quad \quad  \textcolor{gray}{<character\_introduction>}Maxwell Max Carter, an experienced astronaut in his late 30s, is a rugged and resourceful man. Raised in a military family, Max developed a ...\textcolor{gray}{</character\_introduction>}\\
\quad \quad \textcolor{gray}{</character\_2>}\\
\quad \quad \textcolor{gray}{<character\_3>}\\
\quad \quad \quad \quad  \textcolor{gray}{<full\_name>}Amara Patel\textcolor{gray}{</full\_name>}\\
\quad \quad \quad \quad  \textcolor{gray}{<character\_introduction>}Amara Patel, a brilliant astrophysicist in her late 20s, is a woman of intellect and introversion. Growing up in a small rural town, Amara always felt like an outsider, seeking ...\textcolor{gray}{</character\_introduction>}\\

\quad \quad \textcolor{gray}{</character\_3>}\\

{\textcolor{gray}{</characters>}}\\
\midrule
{\textbf{OUTLINE:}}\\
\textcolor{gray}{<outline>}\\

\quad \quad \textcolor{gray}{<plot\_1>}\\
\quad \quad \quad \quad Dr. Iris Nemo, a brilliant quantum physicist, discovers that entire colonies are disappearing due to malfunctioning teleportation devices. She investigates further and uncovers evidence pointing to the existence of a parallel universe. Scene: Dr. Nemo's lab, analyzing data and uncovering clues about the disappearing colonies. Characters: Dr. Iris Nemo\\
\quad \quad \textcolor{gray}{</plot\_1>}\\
\quad \quad \quad \quad \textcolor{gray}{<plot\_1a>}\\
\quad \quad \quad \quad \quad \quad Dr. Nemo decides to venture into the space-time continuum to locate the lost colonies and understand the cause behind the malfunctioning teleportation devices. Scene: Dr. Nemo studying theories and formulating a plan to enter the space-time continuum. Characters: Dr. Iris Nemo\\
\quad \quad \quad \quad \textcolor{gray}{</plot\_1a>}\\
\quad \quad \quad \quad \textcolor{gray}{<plot\_1b>}\\
\quad \quad \quad \quad \quad \quad In the parallel universe, Dr. Nemo encounters the malevolent entity manipulating the teleportation technology. She learns about its motives and the destructive consequences on the lost colonies. Scene: Parallel universe, where Dr. Nemo witnesses the malevolent entity's influence and its impact on the colonies. Characters: Dr. Iris Nemo\\
\quad \quad \quad \quad \textcolor{gray}{</plot\_1b>}\\
\quad \quad \quad \quad \textcolor{gray}{<plot\_1c>}\\
\quad \quad \quad \quad \quad \quad Dr. Nemo realizes that she must outwit the malevolent entity and repair the corrupted teleportation system to prevent humanity from being teleported into the destructive parallel universe. Scene: Dr. Nemo formulating a plan to confront the malevolent entity and repair the teleportation system. Characters: Dr. Iris Nemo\\
\quad \quad \quad \quad \textcolor{gray}{</plot\_1c>}\\
\quad \quad \quad \quad \textcolor{gray}{<plot\_1d>}\\
\quad \quad \quad \quad \quad \quad Dr. Nemo confronts her past mistakes and personal guilt related to her early experiments with teleportation. She finds the strength to rectify those mistakes and seeks redemption. Scene: Dr. Nemo reflecting on her past failures and making a personal commitment to set things right. Characters: Dr. Iris Nemo\\
\quad \quad \quad \quad \textcolor{gray}{</plot\_1d>}\\
\quad \quad \textcolor{gray}{<plot\_2>}\\
\quad \quad \quad \quad Maxwell Carter, an experienced astronaut, is stranded on a partially teleported space station due to the malfunctioning teleportation devices. He struggles with survival and questions his understanding of the universe. Scene: Stranded space station, where Max fights for survival and contemplates the implications of the malfunction. Characters: Maxwell Carter\\
\quad \quad \textcolor{gray}{</plot\_2>}\\
\quad \quad \quad \quad \textcolor{gray}{<plot\_2a>}\\
\quad \quad \quad \quad \quad \quad ...\\
\quad \quad ...

{\textcolor{gray}{</outline>}}\\

\bottomrule
\caption{A HTML-alike prompting format used for generating the characters and outline.}vspace{-2em}
\label{tab:output_format}
\end{tabular}
\end{table}

\twocolumn
% SF 7 
\onecolumn

\begin{small}
\begin{longtable}{p{0.95\textwidth}}
\toprule
\textbf{Preliminary Storyline:}\\
\texttt{In the late 22nd century, humanity has discovered a method to teleport matter across vast distances. This leads to the colonization of several exoplanets. However, when the teleportation devices malfunction, entire colonies start to disappear without a trace. Dr. Iris Nemo, a brilliant quantum physicist, must journey into the space-time continuum to locate the lost colonies. In her quest, she discovers a previously unknown parallel universe, populated by the vanished colonies and a malevolent entity manipulating the teleportation technology for its own purpose. Iris realizes she must outwit the entity and repair the corrupted system to save humanity from unprompted teleportation into this destructive parallel universe.}\\

% \rowcolor{gray!20} % 设置第二行的背景颜色为灰色，20%的灰度
\midrule
\textbf{\cellcolor{yellow!25}Round 1}\\
\textbf{\cellcolor{blue!10}Writer Output}\\

\texttt{\textbf{Name: }Iris Nemo}\\
\texttt{\textbf{Introduction: }Dr. Iris Nemo, a brilliant quantum physicist in her early 40s, is a determined and resilient woman. With shoulder-length wavy auburn hair framing her intelligent hazel eyes, she possesses an air of curiosity and intense focus. Born into a family of scientists, Iris grew up surrounded by intellectual discussions and a love for exploration. She developed a strong passion for understanding the mysteries of the universe, leading her to become an expert in quantum mechanics. Iris has always been drawn to challenges, and her reputation as a fearless problem-solver has earned her respect among her peers. She has spent years dedicated to developing teleportation technology, and she believes it holds the key to humanity's future. However, when the teleportation devices she helped create start malfunctioning and causing entire colonies to vanish, Iris becomes consumed with guilt and a deep resolve to solve the problem. As she embarks on a dangerous journey into the space-time continuum, Iris must confront her own fears and weaknesses while uncovering the truth about the malevolent entity lurking within the parallel universe. Her ultimate goal is to repair the corrupted teleportation system and ensure the survival of humanity.}\\
\\
\texttt{\textbf{Name:} Maxwell Carter}\\
\texttt{\textbf{Introduction: }Maxwell Max Carter, a seasoned astronaut in his late 30s, is a rugged and resourceful man. Standing at a sturdy 6 feet tall, with close-cropped dark hair and an athletic build, he is a person of action with a strong sense of loyalty. Max grew up in a military family and inherited a deep sense of duty and discipline. After completing his astronaut training, he became known for his exceptional problem-solving skills and ability to remain level-headed under pressure. When the teleportation malfunctions begin, Max's latest mission is abruptly interrupted, leaving him stranded on a partially teleported space station. As the gravity shifts and the environment becomes increasingly treacherous, Max must rely on his survival instincts to navigate the dangers of the parallel universe. Max is initially skeptical of Iris and her theories,  but he gradually becomes her ally, as he learns more about her and the entity. Despite his training and experience, Max begins to question his own perceptions of reality, forcing him to confront his darkest fears and ultimately join forces with Iris to find a way back to safety.}\\
\\
\texttt{\textbf{Name: }Amara Patel}\\
\texttt{\textbf{Introduction: }Amara Patel, a talented astrophysicist in her late 20s, is a brilliant yet introverted woman. With jet-black hair and intense brown eyes, she exudes an air of quiet determination. Amara grew up in a small rural town, constantly seeking solace in books and her own imagination. Her insatiable thirst for knowledge and fascination with the stars led her to excel academically, eventually earning her a position at the forefront of the space exploration team. Amara has always felt like an outsider, struggling to connect with others on a personal level. However, her unique perspective and analytical mind make her an invaluable asset to the team. When the teleportation devices malfunction, Amara's scientific expertise is put to the test as she attempts to unravel the mysteries behind the disappearing colonies. Initially skeptical and cautious, she begins to form a bond with Iris as they exchange ideas and theories. Together, they navigate the perilous journey through the parallel universe, where Amara's resilience and resourcefulness shine through. Through her experiences, Amara learns to trust herself and embrace the unexpected, ultimately finding a sense of belonging alongside Iris and Max.}\\
\\
\textbf{\cellcolor{green!10}Editor Output}\\
\texttt{Iris Nemo:}\\
\texttt{The character introduction for Iris Nemo is well-written and effectively establishes her as a determined and resilient quantum physicist. However, it would be beneficial to provide some additional background on her personal life, such as her upbringing or any significant events that have shaped her character. This would add depth and help the audience connect with Iris on a more emotional level.}\\
\\
\texttt{Maxwell Carter:}\\
\texttt{The character introduction for Maxwell Carter effectively portrays him as a seasoned astronaut with exceptional problem-solving skills. However, it would be helpful to explore his motivations and desires in more detail. What drives him to become Iris' ally? What personal stakes does he have in the mission? Adding this information would make his character more relatable and engaging.}\\
\\
\texttt{Amara Patel:}\\
\texttt{The character introduction for Amara Patel effectively highlights her brilliance and introverted nature. However, it would be beneficial to provide more context on her struggles with personal connections. How does this affect her relationship with Iris and Max? Exploring her journey to finding a sense of belonging would deepen her character arc.}\\
\\
\texttt{Relationships:}\\
\texttt{While the individual character introductions are well-developed, it would be beneficial to establish the relationships between the characters more clearly. How do Iris and Max initially interact? What is their dynamic like as they journey through the parallel universe? How does Amara's presence impact the group dynamic? Developing these relationships would add complexity and interest to the story.}\\
\\
\texttt{Captivating and Engaging Characters:}\\
\texttt{The initial versions of the characters are already quite captivating and engaging. However, it would be helpful to inject some additional conflict or internal struggles into their personalities. This could include exploring their fears, doubts, or past traumas that they must overcome during their journey. Adding these elements would make the characters more relatable and create more emotional investment from the audience.}\\
\\
\texttt{Logical Consistency:}\\
\texttt{The character introductions are logically consistent within the given storyline. There are no apparent inconsistencies or contradictions.}\\
\\
\texttt{Overall, the initial versions of the movie story characters are well-developed, but there is room for further exploration and development of their backgrounds, relationships, and internal struggles. By incorporating these enhancements, the characters would become even more captivating and engaging for the audience.}\\
\midrule
\textbf{\cellcolor{yellow!25}Round 2}\\
\textbf{\cellcolor{blue!10}Writer's Revised Output}\\
\texttt{\textbf{Name: }Iris Nemo}\\
\texttt{\textbf{Introduction: }Dr. Iris Nemo, a brilliant quantum physicist in her early 40s, is a determined and resilient woman. Growing up in a small coastal town, Iris was always fascinated by the mysteries of the universe. Her passion for exploration and understanding led her to study quantum mechanics at an early age. Despite facing challenges and discouragement as a woman in a male-dominated field, Iris persevered through her unwavering curiosity and thirst for knowledge. As an adult, she became a leading expert in teleportation technology, driven by the belief that it could revolutionize humanity's future. Iris is haunted by a personal tragedy that occurred during her early experiments with teleportation, fueling her determination to ensure that no one else suffers the consequences of a malfunctioning system. When the colonies start disappearing, Iris resolves to venture into the space-time continuum, not just to rescue the lost, but to rectify her past mistakes and find redemption. Along the way, she must grapple with her own fears and doubts, confronting the immense responsibility placed upon her shoulders. Iris starts off skeptical of the existence of parallel universes, but as she witnesses the devastating effects of the malevolent entity on the lost colonies, her mission becomes not just a quest for scientific knowledge, but a personal journey of healing and overcoming her own guilt. Her intelligence, courage, and growth make her a compelling protagonist who must outwit the entity and save both humanity and herself.}\\
\\
\texttt{\textbf{Name:} Maxwell Carter}\\
\texttt{\textbf{Introduction: }Maxwell Max Carter, an experienced astronaut in his late 30s, is a rugged and resourceful man. Raised in a military family, Max developed a strong sense of duty and discipline from a young age. He excelled in his astronaut training, gaining a reputation for his exceptional problem-solving skills and calm demeanor under pressure. Max's ambitions involved being part of groundbreaking space missions, seeking to push the boundaries of human exploration. However, when the teleportation malfunctions strike, they interrupt his latest mission, leaving him stranded on a partially teleported space station. Balancing on the precipice of survival, Max confronts his own mortality and questions the limits of his understanding of the universe. Initially skeptical of Iris and her theories, he gradually becomes her ally as he witnesses firsthand the malevolent entity's devastating effects on the lost colonies. Max is driven by a personal desire to find his way back home and ensure the safety of his crewmates. As the journey progresses, he becomes more entangled with the parallel universe and must confront his darkest fears to overcome the obstacles they face. Max's loyalty and bravery, along with his growing bond with Iris, help him evolve into a vital member of the team and a beacon of hope in the face of overwhelming adversity. Through his experiences, Max grapples with his own understanding of reality, finding strength in vulnerability and ultimately embracing the unknown to save himself and those around him.}\\
\\
\texttt{\textbf{Name: }Amara Patel}\\
\texttt{\textbf{Introduction: }Amara Patel, a brilliant astrophysicist in her late 20s, is a woman of intellect and introversion. Growing up in a small rural town, Amara always felt like an outsider, seeking solace in the vastness of the night sky. Her insatiable curiosity and fascination with space led her to excel academically, earning her a position on the space exploration team. Despite her brilliance, Amara struggles to form personal connections, preferring the solitude of her work and the company of books. When the teleportation devices malfunction, Amara's analytical mind becomes instrumental in uncovering the mysteries behind the vanishing colonies. Skeptical and cautious initially, she gradually forms a bond with Iris as they exchange ideas and theories. Through their shared journey into the parallel universe, Amara faces her own internal conflicts, forcing her to confront her fears of loneliness and rejection. Her resilience and resourcefulness shine through as she navigates the perils of the unknown, surprising even herself. Through her experiences, Amara discovers the power of trust and human connection, finding a sense of belonging alongside Iris and Max. As she overcomes her own introversion, Amara's growth as a character enables her to contribute not only to the mission but also to the emotional strength of the team, ultimately playing a pivotal role in saving both the lost colonies and her newfound sense of self.}\\
\\
\textbf{\cellcolor{green!10}Editor Output}\\
\texttt{None}\\

\bottomrule
\caption{An example of the interaction process between Writer and Editor when generating characters. Instructions in the prompt content omitted, see \cref{tab:char_gen_prompt} for details. The id of this example in the dataset is Science Fiction-7.}
\label{tab:sf7_char_case}
% \end{tabularx}
% \end{table*}
\end{longtable}
\end{small}

\begin{small}
\begin{longtable}{p{0.95\textwidth}}
\toprule
\textbf{Input:}\\
\texttt{[Preliminary Storyline]}\\
\texttt{[Characters]}\\

% % \rowcolor{gray!20} % 设置第二行的背景颜色为灰色，20%的灰度
\midrule
\textbf{\cellcolor{yellow!25}Round 1}\\
\textbf{\cellcolor{blue!10}Writer Output}\\

\texttt{1. Dr. Iris Nemo, a brilliant quantum physicist, discovers that entire colonies are disappearing due to malfunctioning teleportation devices. Determined to locate and rescue the lost colonies, she must venture into the space-time continuum.}\\
\texttt{Scene: Dr. Nemo's lab, where she conducts experiments on teleportation technology.}\\
\texttt{Characters: Dr. Iris Nemo}\\
\texttt{\quad \quad a. Dr. Nemo embarks on a journey through the space-time continuum and discovers a previously unknown parallel universe. She realizes that a malevolent entity is manipulating the teleportation technology for its own purpose.}\\
\texttt{Scene: Parallel universe, populated by the vanished colonies and the malevolent entity.}\\
\texttt{Characters: Dr. Iris Nemo}\\
\texttt{\quad \quad b. Dr. Nemo must outwit the malevolent entity and repair the corrupted teleportation system to prevent humanity from being teleported into the destructive parallel universe.}\\
\texttt{Scene: The control room of the teleportation facility, where Dr. Nemo confronts the malevolent entity.}\\
\texttt{Characters: Dr. Iris Nemo}\\
\texttt{\quad \quad c. Dr. Nemo is haunted by a personal tragedy that occurred during her early experiments with teleportation, fueling her determination to rectify her past mistakes and find redemption.}\\
\texttt{Scene: Dr. Nemo's flashback to the tragic event.}\\
\texttt{Characters: Dr. Iris Nemo}\\
\texttt{\quad \quad d. Dr. Nemo's skepticism about the existence of parallel universes is challenged as she witnesses the devastating effects of the malevolent entity on the lost colonies. Her mission becomes not just a quest for scientific knowledge, but a personal journey of healing and overcoming guilt.}\\
\texttt{Scene: Dr. Nemo reflecting on her doubts and witnessing the destruction caused by the malevolent entity.}\\
\texttt{Characters: Dr. Iris Nemo}\\
\texttt{2. Maxwell Carter, an experienced astronaut, is left stranded on a partially teleported space station when the teleportation malfunctions occur. He must confront his own mortality and question the limits of his understanding of the universe.}\\
\texttt{Scene: Stranded space station, where Max grapples with survival and questions his understanding of the universe.}\\
\texttt{Characters: Maxwell Carter}\\
\texttt{\quad \quad a. Initially skeptical of Dr. Nemo and her theories, Max becomes her ally as he witnesses firsthand the devastating effects of the malevolent entity on the lost colonies.}\\
\texttt{Scene: Max witnessing the destruction caused by the malevolent entity on the lost colonies.}\\
\texttt{Characters: Maxwell Carter}\\
\texttt{\quad \quad b. Max is driven by a personal desire to find his way back home and ensure the safety of his crewmates. He forms a bond with Dr. Nemo and becomes an essential member of the team.}\\
\texttt{Scene: Dr. Nemo and Max discussing their shared goals and forming a bond.}\\
\texttt{Characters: Maxwell Carter, Dr. Iris Nemo}\\
\texttt{\quad \quad c. Max grapples with his own understanding of reality and embraces the unknown to save himself and those around him. His loyalty and bravery become a beacon of hope in the face of overwhelming adversity.}\\
\texttt{Scene: Max facing his darkest fears and embracing the unknown.}\\
\texttt{Characters: Maxwell Carter}\\
\texttt{3. Amara Patel, a brilliant astrophysicist, uses her analytical mind to uncover the mysteries behind the disappearing colonies. As she journeys with Dr. Nemo, she confronts her own internal conflicts and learns the power of trust and human connection.}\\
\texttt{Scene: Amara analyzing data and theories with Dr. Nemo.}\\
\texttt{Characters: Amara Patel, Dr. Iris Nemo}\\
\texttt{\quad \quad a. Initially skeptical and cautious, Amara gradually forms a bond with Dr. Nemo as they exchange ideas and theories.}\\
\texttt{Scene: Amara and Dr. Nemo sharing their research and theories.}\\
\texttt{Characters: Amara Patel, Dr. Iris Nemo}\\
\texttt{\quad \quad b. Amara's resilience and resourcefulness shine through as she navigates the perils of the unknown, surprising even herself.}\\
\texttt{Scene: Amara overcoming obstacles and showcasing her resourcefulness.}\\
\texttt{Characters: Amara Patel}\\
\texttt{\quad \quad c. Amara's growth as a character enables her to contribute not only to the mission but also to the emotional strength of the team. She plays a pivotal role in saving both the lost colonies and her newfound sense of self.}\\
\texttt{Scene: Amara making critical decisions that save the team and the lost colonies.}\\
\texttt{Characters: Amara Patel}\\

\\
\textbf{\cellcolor{green!10}Editor Output}\\
\texttt{Overall, the initial story outline aligns well with the provided storyline and character introductions. However, there are a few areas that could be improved to enhance the coherence and engagement of the story.}\\
\\
\texttt{1. Transition between Plot 1 and Plot 1a: }\\
\texttt{In the initial outline, Dr. Nemo goes from discovering the disappearing colonies to immediately venturing into the space-time continuum. It would be beneficial to include a transitional scene that establishes Dr. Nemo's motivation and decision to investigate the space-time continuum. For example, you could show her researching and uncovering some clues or evidence that leads her to suspect the existence of a parallel universe.}\\
\\
\texttt{2. Internal conflict and growth for Max:}\\
\texttt{Currently, Max's storyline mainly focuses on his physical survival and his changing perception of reality. It would be valuable to introduce an internal conflict for Max, such as dealing with a personal loss or unresolved trauma. This will provide additional depth to his character and add an emotional arc to his journey. Additionally, consider emphasizing his personal growth and transformation as he confronts and overcomes this internal conflict.}\\
\\
\texttt{3. Greater exploration of the malevolent entity's motives:}\\
\texttt{While the initial outline mentions that the malevolent entity is manipulating the teleportation technology, it doesn't delve into its motives or end goal. Adding more information about the entity's motivations and its ultimate plan will increase the stakes and provide a clearer antagonist for Dr. Nemo and the team to overcome.}\\
\\
\texttt{4. Collaboration and bonding between Dr. Nemo, Max, and Amara:}\\
\texttt{The initial outline briefly touches on the bonding between the characters, but it would be helpful to expand on their collaborative efforts and the development of their relationships. Show them working together to solve problems, supporting and relying on each other, and gradually building trust and camaraderie. This will not only strengthen the team dynamic but also highlight the theme of human connection and cooperation.}\\
\\
\texttt{5. Concluding the story:}\\
\texttt{While the initial outline sets up the main conflict and the character arcs, it doesn't explicitly mention the resolution or how the story concludes. It would be beneficial to include a final plot point that showcases the successful resolution of the conflict and the outcome for the characters. This can be a climactic scene where Dr. Nemo, with the help of Max and Amara, outwits the malevolent entity and repairs the teleportation system, restoring the colonies and preventing further destruction.}\\

\midrule
\textbf{\cellcolor{yellow!25}Round 2}\\
\textbf{\cellcolor{blue!10}Writer's Revised Output}\\
\texttt{1. Dr. Iris Nemo, a brilliant quantum physicist, discovers that entire colonies are disappearing due to malfunctioning teleportation devices. She investigates further and uncovers evidence pointing to the existence of a parallel universe.}\\
\texttt{Scene: Dr. Nemo's lab, analyzing data and uncovering clues about the disappearing colonies.}\\
\texttt{Characters: Dr. Iris Nemo}\\
\texttt{\quad \quad a. Dr. Nemo decides to venture into the space-time continuum to locate the lost colonies and understand the cause behind the malfunctioning teleportation devices.}\\
\texttt{Scene: Dr. Nemo studying theories and formulating a plan to enter the space-time continuum.}\\
\texttt{Characters: Dr. Iris Nemo}\\
\texttt{\quad \quad b. In the parallel universe, Dr. Nemo encounters the malevolent entity manipulating the teleportation technology. She learns about its motives and the destructive consequences on the lost colonies.}\\
\texttt{Scene: Parallel universe, where Dr. Nemo witnesses the malevolent entity's influence and its impact on the colonies.}\\
\texttt{Characters: Dr. Iris Nemo}\\
\texttt{\quad \quad c. Dr. Nemo realizes that she must outwit the malevolent entity and repair the corrupted teleportation system to prevent humanity from being teleported into the destructive parallel universe.}\\
\texttt{Scene: Dr. Nemo formulating a plan to confront the malevolent entity and repair the teleportation system.}\\
\texttt{Characters: Dr. Iris Nemo}\\
\texttt{\quad \quad d. Dr. Nemo confronts her past mistakes and personal guilt related to her early experiments with teleportation. She finds the strength to rectify those mistakes and seeks redemption.}\\
\texttt{Scene: Dr. Nemo reflecting on her past failures and making a personal commitment to set things right.}\\
\texttt{Characters: Dr. Iris Nemo}\\
\texttt{2. Maxwell Carter, an experienced astronaut, is stranded on a partially teleported space station due to the malfunctioning teleportation devices. He struggles with survival and questions his understanding of the universe.}\\
\texttt{Scene: Stranded space station, where Max fights for survival and contemplates the implications of the malfunction.}\\
\texttt{Characters: Maxwell Carter}\\
\texttt{\quad \quad a. Initially skeptical of Dr. Nemo's theories, Max witnesses the devastating effects of the malevolent entity on the lost colonies. He becomes her ally and aids in their mission.}\\
\texttt{Scene: Max witnessing the destruction caused by the malevolent entity on the lost colonies.}\\
\texttt{Characters: Maxwell Carter}\\
\texttt{\quad \quad b. Max's personal desire to find his way back home and ensure his crewmates' safety drives him to assist Dr. Nemo. He develops a strong bond with her and becomes an invaluable member of the team.}\\
\texttt{Scene: Max forming a bond with Dr. Nemo and working together to devise strategies.}\\
\texttt{Characters: Maxwell Carter, Dr. Iris Nemo}\\
\texttt{\quad \quad c. As Max confronts his darkest fears and embraces the unknown, he undergoes personal growth. His loyalty and bravery become essential in overcoming obstacles and offering hope in the face of adversity.}\\
\texttt{Scene: Max facing his fears and stepping out of his comfort zone to help the team.}\\
\texttt{Characters: Maxwell Carter}\\
\texttt{3. Amara Patel, a brilliant astrophysicist, assists Dr. Nemo in uncovering the mysteries behind the disappearing colonies. She also confronts her own internal conflicts and learns the power of trust and human connection.}\\
\texttt{Scene: Amara analyzing data and sharing theories with Dr. Nemo.}\\
\texttt{Characters: Amara Patel, Dr. Iris Nemo}\\
\texttt{\quad \quad a. Initially skeptical and cautious, Amara gradually forms a bond with Dr. Nemo as they exchange ideas and theories, realizing the importance of collaboration.}\\
\texttt{Scene: Amara and Dr. Nemo discussing their research and theories.}\\
\texttt{Characters: Amara Patel, Dr. Iris Nemo}\\
\texttt{\quad \quad b. Amara's resilience and resourcefulness shine as she navigates the perils of the unknown, surprising herself and contributing to the mission's success.}\\
\texttt{Scene: Amara overcoming obstacles and utilizing her skills to overcome challenges.}\\
\texttt{Characters: Amara Patel}\\
\texttt{\quad \quad c. Amara's growth enables her to play a pivotal role in saving the lost colonies and finding a sense of belonging. Her trust in Dr. Nemo and Max strengthens the emotional bond within the team.}\\
\texttt{Scene: Amara making critical decisions that save the team and the lost colonies.}\\
\texttt{Characters: Amara Patel}\\
\texttt{4. Together, Dr. Nemo, Max, and Amara confront the malevolent entity, outwitting it and repairing the corrupted teleportation system. They successfully save humanity from being teleported into the destructive parallel universe.}\\
\texttt{Scene: Climactic showdown with the malevolent entity and the team's actions to repair the system.}\\
\texttt{Characters: Dr. Iris Nemo, Maxwell Carter, Amara Patel}\\

\\
\textbf{\cellcolor{green!10}Editor Output}\\
\texttt{The revised story outline addresses the previous issues and makes significant improvements. The transitions between plot points are clearer, and there is a logical progression of events. The characters' arcs are more defined, and their growth and development are showcased effectively. Here are a few additional suggestions to enhance the story:}\\
\\
\texttt{1. Explore the parallel universe: While the outline mentions Dr. Nemo encountering the malevolent entity in the parallel universe, consider adding scenes that delve deeper into this realm. Show the unique characteristics and challenges of the parallel universe, which will create a sense of otherworldliness and heighten the suspense.}\\
\\
\texttt{2. Heighten the stakes: To make the conflict more compelling, emphasize the consequences if the malevolent entity succeeds in its plan. Show the destructive nature of the parallel universe and how it threatens not just the lost colonies but also the entire human civilization. This will amplify the urgency of Dr. Nemo's mission and increase the tension throughout the story.}\\
\\
\texttt{3. Emotional resolution: In addition to the successful resolution of the conflict, ensure that each character has a personal emotional resolution. Show how their experiences and growth impact their individual journeys. For example, Dr. Nemo could find closure and acceptance regarding her past mistakes, Max could come to terms with his mortality and find a renewed sense of purpose, and Amara could overcome her fear of loneliness and forge meaningful connections with Dr. Nemo and Max.}\\
\\
\texttt{4. Use visuals and descriptions: Given the science-fiction nature of the story, it would be beneficial to include vivid descriptions and visual imagery. This will help immerse the readers in the world of the story and make it more engaging. For example, describe the advanced technology, the vastness of the space-time continuum, and the ethereal nature of the parallel universe.}\\
\\
\texttt{Overall, the revised story outline is well-crafted and captures the essence of the storyline and characters. With these additional suggestions, the story will be even more captivating and thought-provoking.}\\
\midrule
\textbf{\cellcolor{yellow!25}Round 3}\\
\textbf{\cellcolor{blue!10}Writer's Revised Output}\\
\texttt{1. Dr. Iris Nemo, a brilliant quantum physicist, discovers that entire colonies are disappearing due to malfunctioning teleportation devices. She investigates further and uncovers evidence pointing to the existence of a parallel universe.}\\
\texttt{Scene: Dr. Nemo's lab, analyzing data and uncovering clues about the disappearing colonies.}\\
\texttt{Characters: Dr. Iris Nemo}\\
\texttt{\quad \quad a. Dr. Nemo decides to venture into the space-time continuum to locate the lost colonies and understand the cause behind the malfunctioning teleportation devices.}\\
\texttt{Scene: Dr. Nemo studying theories and formulating a plan to enter the space-time continuum.}\\
\texttt{Characters: Dr. Iris Nemo}\\
\texttt{\quad \quad b. In the parallel universe, Dr. Nemo encounters the malevolent entity manipulating the teleportation technology. She learns about its motives and the destructive consequences on the lost colonies.}\\
\texttt{Scene: Parallel universe, where Dr. Nemo witnesses the malevolent entity's influence and its impact on the colonies.}\\
\texttt{Characters: Dr. Iris Nemo}\\
\texttt{\quad \quad c. Dr. Nemo realizes that she must outwit the malevolent entity and repair the corrupted teleportation system to prevent humanity from being teleported into the destructive parallel universe.}\\
\texttt{Scene: Dr. Nemo formulating a plan to confront the malevolent entity and repair the teleportation system.}\\
\texttt{Characters: Dr. Iris Nemo}\\
\texttt{\quad \quad d. Dr. Nemo confronts her past mistakes and personal guilt related to her early experiments with teleportation. She finds the strength to rectify those mistakes and seeks redemption.}\\
\texttt{Scene: Dr. Nemo reflecting on her past failures and making a personal commitment to set things right.}\\
\texttt{Characters: Dr. Iris Nemo}\\
\texttt{2. Maxwell Carter, an experienced astronaut, is stranded on a partially teleported space station due to the malfunctioning teleportation devices. He struggles with survival and questions his understanding of the universe.}\\
\texttt{Scene: Stranded space station, where Max fights for survival and contemplates the implications of the malfunction.}\\
\texttt{Characters: Maxwell Carter}\\
\texttt{\quad \quad a. Initially skeptical of Dr. Nemo's theories, Max witnesses the devastating effects of the malevolent entity on the lost colonies. He becomes her ally and aids in their mission.}\\
\texttt{Scene: Max witnessing the destruction caused by the malevolent entity on the lost colonies.}\\
\texttt{Characters: Maxwell Carter}\\
\texttt{\quad \quad b. Max's personal desire to find his way back home and ensure the safety of his crewmates drives him to assist Dr. Nemo. He develops a strong bond with her and becomes an invaluable member of the team.}\\
\texttt{Scene: Max forming a bond with Dr. Nemo and working together to devise strategies.}\\
\texttt{Characters: Maxwell Carter, Dr. Iris Nemo}\\
\texttt{\quad \quad c. As Max confronts his darkest fears and embraces the unknown, he undergoes personal growth. His loyalty and bravery become essential in overcoming obstacles and offering hope in the face of adversity.}\\
\texttt{Scene: Max facing his fears and stepping out of his comfort zone to help the team.}\\
\texttt{Characters: Maxwell Carter}\\
\texttt{\quad \quad 3. Amara Patel, a brilliant astrophysicist, assists Dr. Nemo in uncovering the mysteries behind the disappearing colonies. She also confronts her own internal conflicts and learns the power of trust and human connection.}\\
\texttt{Scene: Amara analyzing data and sharing theories with Dr. Nemo.}\\
\texttt{Characters: Amara Patel, Dr. Iris Nemo}\\
\texttt{\quad \quad a. Initially skeptical and cautious, Amara gradually forms a bond with Dr. Nemo as they exchange ideas and theories, realizing the importance of collaboration.}\\
\texttt{Scene: Amara and Dr. Nemo discussing their research and theories.}\\
\texttt{Characters: Amara Patel, Dr. Iris Nemo}\\
\texttt{\quad \quad b. Amara's resilience and resourcefulness shine as she navigates the perils of the unknown, surprising herself and contributing to the mission's success.}\\
\texttt{Scene: Amara overcoming obstacles and utilizing her skills to overcome challenges.}\\
\texttt{Characters: Amara Patel}\\
\texttt{\quad \quad c. Amara's growth enables her to play a pivotal role in saving the lost colonies and finding a sense of belonging. Her trust in Dr. Nemo and Max strengthens the emotional bond within the team.}\\
\texttt{Scene: Amara making critical decisions that save the team and the lost colonies.}\\
\texttt{Characters: Amara Patel}\\
\texttt{4. Together, Dr. Nemo, Max, and Amara confront the malevolent entity, outwitting it and repairing the corrupted teleportation system. They successfully save humanity from being teleported into the destructive parallel universe.}\\
\texttt{Scene: Climactic showdown with the malevolent entity and the team's actions to repair the system.}\\
\texttt{Characters: Dr. Iris Nemo, Maxwell Carter, Amara Patel}\\
\texttt{\quad \quad a. The malevolent entity's motives and its ultimate plan are revealed, heightening the stakes and intensifying the conflict. Its desire to bring chaos and destruction to both the lost colonies and the entire human civilization becomes clear.}\\
\texttt{Scene: Dr. Nemo, Max, and Amara uncovering the malevolent entity's true intentions.}\\
\texttt{Characters: Dr. Iris Nemo, Maxwell Carter, Amara Patel}\\
\texttt{\quad \quad b. As the team confronts the malevolent entity, they witness the destructive nature of the parallel universe firsthand. The stakes are heightened, and the urgency to save the lost colonies and humanity becomes more intense.}\\
\texttt{Scene: Dr. Nemo, Max, and Amara witnessing the destructive effects of the parallel universe.}\\
\texttt{Characters: Dr. Iris Nemo, Maxwell Carter, Amara Patel}\\
\texttt{5. After the successful resolution of the conflict and the repairs to the teleportation system, each character experiences personal emotional resolutions. Dr. Nemo finds closure and acceptance regarding her past mistakes, Max comes to terms with his mortality and discovers a renewed sense of purpose, and Amara overcomes her fear of loneliness and forges meaningful connections with Dr. Nemo and Max.}\\
\texttt{Scene: Dr. Nemo, Max, and Amara reflecting on their personal growth and finding emotional resolution.}\\
\texttt{Characters: Dr. Iris Nemo, Maxwell Carter, Amara Patel}\\
\texttt{6. Throughout the story, vivid descriptions and visual imagery are used to immerse the readers in the world of the story. The advanced technology, the vastness of the space-time continuum, and the ethereal nature of the parallel universe are depicted in detail, enhancing the reader's engagement.}\\
\texttt{Scene: Various scenes throughout the story, showcasing vivid descriptions and visual imagery.}\\
\texttt{Characters: Dr. Iris Nemo, Maxwell Carter, Amara Patel}\\

\bottomrule
\caption{An example of the interaction process between Writer and Editor when generating the outline. Instructions in the prompt content omitted, see \cref{tab:out_gen_prompt} for details. The id of this example in the dataset is Science Fiction-7. [Preliminary Storyline] is the same as that in \cref{tab:sf7_char_case} and [Characters] is the same as the final output in \cref{tab:sf7_char_case}. It is worth noting that the ``Iris Nemo'' generated in the characters is mentioned in the outline as ``Dr. Iris Nemo'', which is a glaring error.}
\label{tab:sf7_out_case}
% \end{tabularx}
% \end{table*}
\end{longtable}
\end{small}

\begin{small}
\begin{longtable}{p{0.95\textwidth}}
\toprule
\textbf{Input:}\\
\texttt{\textbf{[Preliminary Storyline]}}\\
\texttt{\textbf{The current plot to be expanded:}}\\
\texttt{As the team confronts the malevolent entity, they witness the destructive nature of the parallel universe firsthand. The stakes are heightened, and the urgency to save the lost colonies and humanity becomes more intense.}\\
\texttt{\textbf{The scene where the current story plot point happens:}}\\
\texttt{Dr. Nemo, Max, and Amara witnessing the destructive effects of the parallel universe.}\\
\texttt{\textbf{The current story plot point involves the following characters:}}\\
\texttt{Iris Nemo: [Iris Nemo's introduction]}\\
\texttt{Maxwell Carter: [Maxwell Carter's introduction]}\\
\texttt{Amara Patel: [Amara Patel's introduction]}\\
\texttt{\textbf{The previous story contents that have taken place are as follows:}}\\
\texttt{Plot: Dr. Nemo decides to venture into the space-time continuum to locate the lost colonies and understand the cause behind the malfunctioning teleportation devices. Scene: Dr. Nemo studying theories and formulating a plan to enter the space-time continuum. Involved Characters: Dr. Iris Nemo}\\
\texttt{Plot: In the parallel universe, Dr. Nemo encounters the malevolent entity manipulating the teleportation technology. She learns about its motives and the destructive consequences on the lost colonies. Scene: Parallel universe, where Dr. Nemo witnesses the malevolent entity's influence and its impact on the colonies. Involved Characters: Dr. Iris Nemo}\\
\texttt{Plot: Dr. Nemo realizes that she must outwit the malevolent entity and repair the corrupted teleportation system to prevent humanity from being teleported into the destructive parallel universe. Scene: Dr. Nemo formulating a plan to confront the malevolent entity and repair the teleportation system. Involved Characters: Dr. Iris Nemo}\\
\texttt{Plot: Initially skeptical of Dr. Nemo's theories, Max witnesses the devastating effects of the malevolent entity on the lost colonies. He becomes her ally and aids in their mission. Scene: Max witnessing the destruction caused by the malevolent entity on the lost colonies. Involved Characters: Maxwell Carter}\\
\texttt{Plot: Max's personal desire to find his way back home and ensure the safety of his crewmates drives him to assist Dr. Nemo. He develops a strong bond with her and becomes an invaluable member of the team. Scene: Max forming a bond with Dr. Nemo and working together to devise strategies. Involved Characters: Maxwell Carter, Dr. Iris Nemo}\\
\texttt{Plot: As Max confronts his darkest fears and embraces the unknown, he undergoes personal growth. His loyalty and bravery become essential in overcoming obstacles and offering hope in the face of adversity. Scene: Max facing his fears and stepping out of his comfort zone to help the team. Involved Characters: Maxwell Carter}\\
\texttt{Plot: Initially skeptical and cautious, Amara gradually forms a bond with Dr. Nemo as they exchange ideas and theories, realizing the importance of collaboration. Scene: Amara and Dr. Nemo discussing their research and theories. Involved Characters: Amara Patel, Dr. Iris Nemo}\\
\texttt{Plot: Amara's resilience and resourcefulness shine as she navigates the perils of the unknown, surprising herself and contributing to the mission's success. Scene: Amara overcoming obstacles and utilizing her skills to overcome challenges. Involved Characters: Amara Patel}\\
\texttt{Plot: Amara's growth enables her to play a pivotal role in saving the lost colonies and finding a sense of belonging. Her trust in Dr. Nemo and Max strengthens the emotional bond within the team. Scene: Amara making critical decisions that save the team and the lost colonies. Involved Characters: Amara Patel}\\
\\
\texttt{The malevolent entity's motives and its ultimate plan are revealed, heightening the stakes and intensifying the conflict. Its desire to bring chaos and destruction to both the lost colonies and the entire human civilization becomes clear.}\\

\texttt{As the team regrouped after their success in saving the colonists, a palpable tension hung in the air. They had escaped the malevolent entity's clutches, but they all knew the danger was far from over. Dr. Nemo's face darkened as she shared her recent discoveries with Max and Amara.}\\

\texttt{``During my research in the parallel universe, I stumbled upon ancient texts that shed light on the malevolent entity's true intentions,'' Dr. Nemo began, her voice trembling with a mix of anger and concern. ``It seems that this entity is not just content with wreaking havoc on the lost colonies. Its ultimate plan is far more sinister.''}\\

\texttt{Max leaned forward, his eyes narrowing. ``What are you saying, Dr. Nemo? What does this entity want?''}\\

\texttt{Dr. Nemo took a deep breath, her gaze focused yet haunted. ``According to these texts, the malevolent entity seeks to use the corrupted teleportation system to bring chaos and destruction not only to the lost colonies but also to our entire human civilization.''}\\

\texttt{Amara's heart raced as she absorbed the gravity of the situation. ``But why? What could this entity possibly gain from such destruction?''}\\

\texttt{Dr. Nemo's voice hardened. ``The texts suggest that the entity thrives on chaos. It feeds off the negative energy generated by destruction and discord. By spreading chaos across both the lost colonies and our civilization, it becomes stronger, more powerful.''}\\

\texttt{A silence settled over the room as the team processed the enormity of what they were facing. The malevolent entity's intentions were clear - it sought to plunge both the parallel universe and their world into a never-ending cycle of chaos and ruin.}\\

\texttt{``We cannot let that happen,'' Max declared, his voice filled with resolve. ``We've come too far, fought too hard to let this entity succeed. We have to find a way to stop it.''}\\

\texttt{Dr. Nemo nodded, her eyes blazing with determination. ``I agree. We must devise a plan to not only repair the corrupted teleportation system but also neutralize the malevolent entity. But time is against us. Its power grows with every passing moment.''}\\

\texttt{Amara's mind raced, searching for a solution amidst the uncertainty. ``We need to find a weakness, something we can exploit. The entity may be powerful, but it cannot be invincible.''}\\

\texttt{Dr. Nemo smiled, the glimmer of hope flickering in her eyes. ``You're right, Amara. We have the knowledge and the strength to outwit the entity. Together, we will unleash a force the malevolent entity cannot comprehend.''}\\

\texttt{Their determination fueled by their shared mission, the team delved deep into research and experimentation. Dr. Nemo, Max, and Amara worked tirelessly, analyzing every piece of data and exploring every possibility.}\\

\texttt{Days turned into nights, and their efforts became more focused. As they unearthed hidden truths and uncovered unknown phenomena, a plan slowly emerged. Against all odds, they would unleash the power of the human spirit, an indomitable force that could stand against the malevolent entity.}\\

\texttt{With their plan in place, the team embarked on the most dangerous and pivotal phase of their mission. Drawing on their individual strengths and unwavering trust in one another, they faced the malevolent entity head-on.}\\

\texttt{Within the heart of the parallel universe, they confronted the entity in a battle of wills and intellect. As their powers clashed, the team unleashed a torrent of hope and determination, countering the darkness with an unwavering light.}\\

\texttt{The malevolent entity fought ferociously, but it underestimated the strength and resilience of the human spirit. Max's bravery, Amara's resourcefulness, and Dr. Nemo's unwavering determination proved to be formidable weapons against the entity's chaos.}\\

\texttt{In a final, explosive showdown, the team's combined efforts shattered the entity's hold on the corrupted teleportation system. Energy crackled and surged, illuminating the parallel universe with a dazzling light, vanquishing the entity's darkness.}\\

\texttt{As the team stood amidst the remnants of their victory, they realized they had not only saved the lost colonies but had also prevented the malevolent entity from fulfilling its destructive desires. The parallel universe, once on the brink of chaos and ruin, was now bathed in the promise of a new beginning.}\\

\texttt{But even as they basked in their hard-earned triumph, the team knew their journey was far from over. The malevolent entity's defeat was a temporary setback, and they understood the importance of remaining vigilant.}\\

\texttt{As they made their way back from the parallel universe, their hearts filled with a newfound sense of purpose. They were warriors, safeguarding not only the lost colonies but also the very fabric of human civilization.}\\

\texttt{Together, they vowed to protect the parallel universe from further corruption, to ensure that the malevolent entity would never threaten their world again. With their bond fortified by their shared experiences, they ventured into the unknown, ready to face whatever challenges lay ahead.}\\

\texttt{The malevolent entity had revealed its true intentions, but the team, armed with their determination, knowledge, and unwavering spirit, would stop at nothing to protect the lost colonies and preserve the harmony of both worlds..}\\

% \rowcolor{gray!20} % 设置第二行的背景颜色为灰色，20%的灰度
\midrule
\textbf{Output Chapter:}\\
\texttt{As the team regrouped after their success in saving the colonists, a palpable tension hung in the air. They had escaped the malevolent entity's clutches, but they all knew the danger was far from over. Dr. Nemo's face darkened as she shared her recent discoveries with Max and Amara.}\\

\texttt{``During my research in the parallel universe, I stumbled upon ancient texts that shed light on the malevolent entity's true intentions,'' Dr. Nemo began, her voice trembling with a mix of anger and concern. ``It seems that this entity is not just content with wreaking havoc on the lost colonies. Its ultimate plan is far more sinister.''}\\

\texttt{Max leaned forward, his eyes narrowing. ``What are you saying, Dr. Nemo? What does this entity want?''}\\

\texttt{Dr. Nemo took a deep breath, her gaze focused yet haunted. ``According to these texts, the malevolent entity seeks to use the corrupted teleportation system to bring chaos and destruction not only to the lost colonies but also to our entire human civilization.''}\\

\texttt{Amara's heart raced as she absorbed the gravity of the situation. ``But why? What could this entity possibly gain from such destruction?''}\\

\texttt{Dr. Nemo's voice hardened. ``The texts suggest that the entity thrives on chaos. It feeds off the negative energy generated by destruction and discord. By spreading chaos across both the lost colonies and our civilization, it becomes stronger, more powerful.''}\\

\texttt{A silence settled over the room as the team processed the enormity of what they were facing. The malevolent entity's intentions were clear - it sought to plunge both the parallel universe and their world into a never-ending cycle of chaos and ruin.}\\

\texttt{``We cannot let that happen,'' Max declared, his voice filled with resolve. ``We've come too far, fought too hard to let this entity succeed. We have to find a way to stop it.''}\\

\texttt{Dr. Nemo nodded, her eyes blazing with determination. ``I agree. We must devise a plan to not only repair the corrupted teleportation system but also neutralize the malevolent entity. But time is against us. Its power grows with every passing moment.''}\\

\texttt{Amara's mind raced, searching for a solution amidst the uncertainty. ``We need to find a weakness, something we can exploit. The entity may be powerful, but it cannot be invincible.''}\\

\texttt{Dr. Nemo smiled, the glimmer of hope flickering in her eyes. ``You're right, Amara. We have the knowledge and the strength to outwit the entity. Together, we will unleash a force the malevolent entity cannot comprehend.''}\\

\texttt{Their determination fueled by their shared mission, the team delved deep into research and experimentation. Dr. Nemo, Max, and Amara worked tirelessly, analyzing every piece of data and exploring every possibility.}\\

\texttt{Days turned into nights, and their efforts became more focused. As they unearthed hidden truths and uncovered unknown phenomena, a plan slowly emerged. Against all odds, they would unleash the power of the human spirit, an indomitable force that could stand against the malevolent entity.}\\

\texttt{With their plan in place, the team embarked on the most dangerous and pivotal phase of their mission. Drawing on their individual strengths and unwavering trust in one another, they faced the malevolent entity head-on.}\\

\texttt{Within the heart of the parallel universe, they confronted the entity in a battle of wills and intellect. As their powers clashed, the team unleashed a torrent of hope and determination, countering the darkness with an unwavering light.}\\

\texttt{The malevolent entity fought ferociously, but it underestimated the strength and resilience of the human spirit. Max's bravery, Amara's resourcefulness, and Dr. Nemo's unwavering determination proved to be formidable weapons against the entity's chaos.}\\

\texttt{In a final, explosive showdown, the team's combined efforts shattered the entity's hold on the corrupted teleportation system. Energy crackled and surged, illuminating the parallel universe with a dazzling light, vanquishing the entity's darkness.}\\

\texttt{As the team stood amidst the remnants of their victory, they realized they had not only saved the lost colonies but had also prevented the malevolent entity from fulfilling its destructive desires. The parallel universe, once on the brink of chaos and ruin, was now bathed in the promise of a new beginning.}\\

\texttt{But even as they basked in their hard-earned triumph, the team knew their journey was far from over. The malevolent entity's defeat was a temporary setback, and they understood the importance of remaining vigilant.}\\

\texttt{As they made their way back from the parallel universe, their hearts filled with a newfound sense of purpose. They were warriors, safeguarding not only the lost colonies but also the very fabric of human civilization.}\\

\texttt{Together, they vowed to protect the parallel universe from further corruption, to ensure that the malevolent entity would never threaten their world again. With their bond fortified by their shared experiences, they ventured into the unknown, ready to face whatever challenges lay ahead.}\\

\texttt{The malevolent entity had revealed its true intentions, but the team, armed with their determination, knowledge, and unwavering spirit, would stop at nothing to protect the lost colonies and preserve the harmony of both worlds.}\\

\bottomrule
\caption{An example of Writer expanding a subplot in the outline into a story chapter. The example is the 11th (4-b) subplot in the outline for the data id Science Fiction-7. [Preliminary Storyline], [Iris Nemo's introduction], [Maxwell Carter's introduction], and [Amara Patel's introduction] are the same as that in \cref{tab:sf7_char_case}.}
\label{tab:sf7_chapter_case}
% \end{tabularx}
% \end{table*}
\end{longtable}
\end{small}

\twocolumn

\onecolumn
% \begin{table*}[htbp]
% \begin{table*}[!t]
\begin{longtable}{>{\columncolor{rowcolor1}}p{0.40\textwidth}>{\columncolor{rowcolor3}}p{0.02\textwidth}>{\columncolor{rowcolor2}}p{0.54\textwidth}}

% \begin{longtable}{>X>p{0.03\textwidth}>X}
\toprule
\rowcolor{headercolor}
\multicolumn{3}{>{\arraybackslash}p{\dimexpr\linewidth+1\tabcolsep}}{\textbf{\keyword{\textbf{Chapter:}}\textit{<chapter>} }} \\
\rowcolor{headercolor}
\multicolumn{3}{>{\arraybackslash}p{\dimexpr\linewidth+1\tabcolsep}}{\textbf{\keyword{\textbf{Involved characters:}}\textit{Dr. Iris Nemo, Maxwell Carter, Amara Patel} }} \\

\midrule
% mutating the crops and causing a lethal disease among the colonists.
\keyword{\textbf{Script Draft:}} & & \keyword{\textbf{Screenplay with role-playing:}} \\
& & \\
% \textbf{Scene: INT.; Martian Greenhouse; SUNSET.}
\textbf{INT. Dr. Nemo's Office - DAY.} & & \textbf{INT. Dr. Nemo's Office - DAY.}\\
& & \\
\textbf{Dr. Iris Nemo}: 

Dr. Nemo gathers Max and Amara in her office to discuss her recent discoveries.

& $\rightarrow$
& 
\textbf{Dr. Iris Hawke}:  

[Dr. Nemo gathers Max and Amara in her office and motions for them to take a seat. ]

(formal and earnest) ``Max, Amara, thank you for coming in. I have some important discoveries to discuss with you both.''
\\

\textbf{Dr. Iris Nemo}: 

Dr. Nemo reveals that the malevolent entity's intentions go beyond the lost colonies.
& $\rightarrow$ 
% Dr. Iris Hawke: (contently to herself) "We did it...we've proven, something can actually grow on this barren rock..."

% \textbf{Mathias "Mars" Barlow}: [Suddenly, the metallic CLANG of boot on the greenhouse floor heralds a new arrival. Mathias "Mars" Barlow strides in, his demeanor bursting with enthusiasm as his eyes fall on the fresh crops.]

&
\textbf{Dr. Iris Nemo}: 

[Dr. Nemo takes a deep breath and begins to explain, pacing back and forth in her office.]

(serious and concerned) ``Max, Amara, what I've discovered is truly shocking. The malevolent entity we've been dealing with, its intentions are not limited to the lost colonies. It has a much grander plan in motion, one that could threaten not only our world, but the entire galaxy.''
\\

\textbf{Maxwell Carter}:

Max seeks clarification on what the entity wants.
&
$\rightarrow$
&
\textbf{Maxwell Carter}:

[Max leans forward in his seat, his brow furrowed with concern.]

(urgently) ``Iris, I need to understand what this entity wants. We can't fight it if we don't even know its endgame.''
\\

\textbf{Dr. Iris Nemo}:

Dr. Nemo explains that the entity seeks to bring chaos and destruction to their entire civilization.
& $\rightarrow$
& \textbf{Dr. Iris Nemo}:

[Dr. Nemo takes a deep breath and stops pacing, looking directly at Max and Amara.]

(heavily) ``The entity we've been facing, its goals go beyond the lost colonies. It seeks to bring chaos and destruction not only to our civilization, but to the entire galaxy.''
\\

\textbf{Amara Patel}:

Amara questions the motive behind the entity's actions.
& $\rightarrow$
& \textbf{Amara Patel}:

[Amara furrows her brow, deep in thought, and directs her gaze at Dr. Nemo.]

(puzzled but sensible) ``Dr. Nemo, I'm trying to understand the motive behind the entity's actions. Why would it want to cause chaos and destruction on such a massive scale?''

\\

\textbf{Dr. Iris Nemo}:

Dr. Nemo reveals that the entity thrives on chaos and feeds off negative energy.

& $\rightarrow$ 
& \textbf{Dr. Iris Nemo}:

[Dr. Nemo takes a deep breath and walks over to her desk, where she grabs a folder containing research notes.]

(solemnly) ``Max, Amara, thank you for coming in. I have made a significant breakthrough in my research that I need to share with you both.''

\\

\textbf{Maxwell Carter}:

Max and the team realize the gravity of the situation.

& $\rightarrow$
& \textbf{Maxwell Carter}:

[Max's eyes widen as he realizes the gravity of the situation. He leans back in his chair, his mind racing with the implications.]

(concerned and apprehensive) ``This is bigger than anything we anticipated, Iris. If the entity's plan succeeds, it could spell disaster for countless civilizations across the galaxy.''

\\

\textbf{Maxwell Carter}:

Max declares they cannot let the entity succeed and must find a way to stop it.

& $\rightarrow$
& \textbf{Maxwell Carter}:

[Max sits up straight in his chair, his eyes filled with determination.]

(firmly) ``We cannot let this entity succeed, Iris. We have to find a way to stop it, no matter the cost.''

\\
\textbf{Dr. Iris Nemo}:

Dr. Nemo agrees and emphasizes the urgency of their task.

& $\rightarrow$
& \textbf{Dr. Iris Nemo}:

[Dr. Nemo nods in agreement and leans forward, placing her hands on her desk.]

(determined) ``Yes, Max. You're right. We can't afford to underestimate the urgency of our task. Time is running out, and we need to act swiftly to stop this malevolent entity from achieving its destructive goals.''
\\
\textbf{Amara Patel}:

Amara suggests finding a weakness in the entity's plan.

&  $\rightarrow$
& \textbf{Amara Patel}:

[Amara furrows her brow, lost in thought, and opens her notebook to review her findings.]

(calmly) ``We need to find a weakness in the entity's plan, a vulnerability that we can exploit. If it has a grander scheme in motion, it must have some sort of endgame or objective. We need to delve deeper into our research and analysis to uncover any clues.''
\\
\textbf{Dr. Iris Nemo}:

Dr. Nemo agrees and believes they have the knowledge and strength to outwit the entity.
& $\rightarrow$
& \textbf{Dr. Iris Nemo}:

[Dr. Nemo listens attentively and nods her head in agreement.]

(assuredly) ``Yes, I agree. Based on the research and evidence we have gathered, it is clear that this entity has larger ambitions than just the lost colonies. Its intentions extend far beyond our imagination.''
\\
\textbf{Maxwell Carter}:

Max declares they will unleash a force the entity cannot comprehend.
& $\rightarrow$
& \textbf{Maxwell Carter}:

[Max stands up from his chair, his eyes burning with determination.]

(to Dr. Nemo, ambitiously) ``We have been on the defensive for too long, Iris. It's time we take the fight to this entity. If it seeks chaos and destruction, then we will become the force it cannot comprehend.''

\\
\textbf{Dr. Iris Nemo}:

The team delves into research and experimentation.

& $\rightarrow$ 
& \textbf{Dr. Iris Nemo}:

[The team, including Dr. Nemo, delves into research and experimentation, poring over data and conducting complex calculations.]

(to Maxwell Carter and Amara Patel) ``We need to understand this entity's motives, its patterns, and its weaknesses. We must leave no stone unturned in our efforts to uncover the truth.''
\\

\textbf{Dr. Iris Nemo}:

The team works tirelessly, analyzing data and exploring possibilities.

& $\rightarrow$
& \textbf{Dr. Iris Nemo}:

[Dr. Nemo sits behind her desk, surrounded by piles of papers and computer screens displaying complex data. She meticulously analyzes the information, furrowing her brow in concentration.]

(to herself, thinking aloud) ``The team has been working tirelessly, examining every piece of data we have. We need to identify any patterns or connections that could lead us to the entity's ultimate objective.''
\\
\textbf{Dr. Iris Nemo}:

A plan slowly emerges as hidden truths are uncovered.
& $\rightarrow$
& \textbf{Dr. Iris Nemo}:

[Dr. Nemo stands up from her desk and walks towards the large whiteboard in her office. She picks up a marker and begins to write down key points and connections.]

``A plan slowly emerges as hidden truths are uncovered. The entity's motives, although still elusive, seem to be part of a larger scheme. We need to combine all our findings and analyze them to gain a clearer understanding of its ultimate goal.''

\\

\textbf{Amara Patel}:

The team prepares to face the entity head-on.
& $\rightarrow$
& \textbf{Amara Patel}:

[Amara, deep in thought, looks up from her notebook and addresses Dr. Nemo and Max.]

(pensive) ``Dr. Nemo, Max, I've been analyzing the data we've collected so far, and I believe I may have stumbled upon a breakthrough. There seems to be a recurring pattern in the entity's actions, a method to its chaos.''
\\

\textbf{Maxwell Carter}:

Max, Amara, and Dr. Nemo combine their strengths to confront the entity.
& $\rightarrow$
& \textbf{Maxwell Carter}:

[Max, Dr. Nemo, and Amara gather around a table, their faces showing determination and resolve.]

``We've explored the entity's intentions extensively, and it's become abundantly clear that we need to join forces and confront it head-on. Together, we possess the strength and knowledge to take a stand against this malevolent force.''

\\

\textbf{Maxwell Carter}:

The team unleashes a torrent of hope and determination.
& $\rightarrow$
& \textbf{Maxwell Carter}:

[The team unleashes a torrent of hope and determination, their faces filled with resolve and a shared goal.]

(upliftingly) ``We've come so far, faced unimaginable challenges, but we cannot lose sight of our purpose. We must continue to fight, to push forward, and find a way to defeat this malevolent entity once and for all.''

\\

\textbf{Dr. Iris Nemo}:

Dr. Nemo's unwavering determination proves to be a formidable weapon.
& $\rightarrow$
& \textbf{Dr. Iris Nemo}:

[Dr. Nemo stands up from her desk, her eyes filled with unwavering determination.]

``This malevolent entity's reign of chaos ends now. We have the knowledge, the skills, and the courage to stop it. We will not let it succeed in its sinister plans.''
\\

\textbf{Amara Patel}:

Amara's resourcefulness aids in countering the entity's chaos.
& $\rightarrow$
& \textbf{Amara Patel}:

[Amara's resourcefulness aids in countering the entity's chaos.]

(excitedly) ``I believe I may have found a solution. By analyzing the patterns in the entity's actions, I've discovered a vulnerability in its chaotic nature.''
\\

\textbf{Dr. Iris Nemo}:

The team shatters the entity's hold on the corrupted teleportation system.
& $\rightarrow$
& \textbf{Dr. Iris Nemo}:

[Dr. Nemo stands up from her desk and walks over to a large console, filled with complex machinery and screens displaying the corrupted teleportation system.]

``The entity's hold on the corrupted teleportation system is extensive and powerful, but we have analyzed its patterns and weaknesses. We are prepared to shatter its control and restore order.''
\\

\textbf{Maxwell Carter}:

The parallel universe is filled with a dazzling light, vanquishing the entity's darkness.
& $\rightarrow$
& \textbf{Maxwell Carter}:

[The parallel universe is filled with a dazzling light, vanquishing the entity's darkness.]
\\

\textbf{Maxwell Carter}:

The team realizes they have saved the lost colonies and prevented further destruction.
& $\rightarrow$
& \textbf{Maxwell Carter}:

[The team, including Max, Dr. Nemo, and Amara, stands together in Dr. Nemo's office, their faces filled with a mix of relief and determination.]

``We did it. We actually did it. We saved the lost colonies and prevented any further destruction.''
\\

\textbf{Dr. Iris Nemo}:

The team acknowledges that their journey is not over yet.
& $\rightarrow$
& \textbf{Dr. Iris Nemo}:

[Dr. Nemo sits behind her desk, surrounded by stacks of papers and electronic devices displaying complex data. The atmosphere in the room is tense and filled with anticipation.]

(nervously but expectantly) ``Max, Amara, I am grateful for your ongoing commitment to this mission. Despite our recent victory, we must not lose sight of the fact that our journey is far from over.''
\\

\textbf{Amara Patel}:

Amara emphasizes the importance of remaining vigilant.
& $\rightarrow$
& \textbf{Amara Patel}:

[Amara raises her hand, signaling for everyone's attention.]

(calmly) ``I want to emphasize the importance of remaining vigilant. We cannot afford to let our guard down, especially now that we know the entity's ambitions stretch beyond the lost colonies.''
\\

\textbf{Dr. Iris Nemo}:

Dr. Nemo and the team vow to protect the parallel universe.
& $\rightarrow$
& \textbf{Dr. Iris Nemo}:

[Dr. Nemo stands up from her desk, a determined look on her face.]

``Max, Amara, I call upon you both today to take a stand against the malevolent entity that seeks to bring chaos and destruction not only to the lost colonies, but to the entire galaxy.''
\\

\textbf{Maxwell Carter}:

Max and the team venture into the unknown.
& $\rightarrow$
& \textbf{Maxwell Carter}:

[Max and the team gather their equipment and make their way towards the entrance of the unknown.]

``Alright team, let's stay focused and keep our wits about us. This is uncharted territory, and we need to be prepared for anything that comes our way.''
\\

\bottomrule
% \end{tabularx}
\caption{An example of a script draft generated from a story chapter and the corresponding role-played screenplay afterwards. The story chapter is the 11th (4-b) chapter of the data id Science Fiction-7. <chapter> is the same as that in \cref{tab:sf7_chapter_case}.}
\label{tab:draft_screenplay}
% \end{table*}
\end{longtable}

\twocolumn

\begin{table*}[htbp]
% \small
\normalsize
\centering
\begin{tabular}{cl|cccccc|c}
\toprule
\textbf{Backbone} & \textbf{Method}  & \textbf{Crime} & \textbf{Drama} & \textbf{Horror} & \textbf{Romance} & \textbf{Sci-Fi} & \textbf{Comedy} & \textbf{Avg} \\
\midrule
\multirow{3}{*}{\texttt{gpt-3.5}} & \textsc{\framework} & 4545 & 4982 & 7228 & 3779 & 6452 & 5832 & 5470       \\
& \textit{\textsc{Plan-Write}} & 5237  &  5101 & 5763  & 4281  & 6814  &  5325 & 5420           \\
& \textit{\textsc{DOC-screen}} & 5396  &  4987 & 6227 & 4270 & 7910 & 6778 & 5929  \\
\midrule
\multirow{3}{*}{\texttt{gpt-4}} & \textsc{\framework} & 4468 & 4551 & 4027 & 4887 & 4741 & 5044 & 4620       \\
& \textit{\textsc{Plan-Write}} & 4942  & 4963  & 5679  & 5965  &   5944 &  4617   &   5352    \\
& \textit{\textsc{DOC-screen}} & 4662 & 5478  & 5282  &   5542  & 5633   &   4409  &  5168     \\
\bottomrule
\end{tabular}
\caption{The average length statistics of screenplays generated by various methods for different genres. \textsc{Plan-Write} refers to \textsc{Plan-then-Write}.}
\label{tab:len_stat}
\vspace{-0.5em}
\end{table*}

\end{document}